\documentclass{article}

\usepackage[preprint]{neurips_2025}

\usepackage[utf8]{inputenc} 
\usepackage[T1]{fontenc}    
\usepackage{hyperref}       
\usepackage{url}            
\usepackage{booktabs}       
\usepackage{amsfonts}       
\usepackage{nicefrac}       
\usepackage{microtype}      

\usepackage{amsmath}   
\usepackage{algorithm}
\usepackage{algorithmicx}
\usepackage{algpseudocode}
\usepackage{graphicx}
\usepackage[table,xcdraw]{xcolor}
\usepackage[most]{tcolorbox}
\usepackage{enumitem}

\title{SHARP: Synthesizing High-quality Aligned Reasoning Problems for Large Reasoning Models Reinforcement Learning}

\author{
 Xiong Jun Wu\textsuperscript{*1} \hspace{3mm} Zhenduo Zhang\textsuperscript{*1} \hspace{3mm} Zujie Wen\textsuperscript{1} \hspace{3mm} Zhiqiang Zhang\textsuperscript{1} \hspace{3mm} Wang Ren\textsuperscript{2} \\
\textbf{Lei Shi}\textsuperscript{2} \hspace{3mm} \textbf{Cai Chen}\textsuperscript{2}  \hspace{3mm} \textbf{Deng Zhao}\textsuperscript{2} \hspace{3mm} \textbf{Qing Wang}\textsuperscript{3} \hspace{3mm} \textbf{Xudong Han}\textsuperscript{3} \hspace{3mm} \textbf{Chengfu Tang}\textsuperscript{3} \\
\textbf{Dingnan Jin}\textsuperscript{2} \hspace{3mm} \textbf{Qing Cui}\textsuperscript{2} \hspace{3mm} \textbf{Jun Zhou}\textsuperscript{1}\\
\textsuperscript{1} AI Alignment at Ant Group \hspace{4mm} \textsuperscript{1,2} NextEvo, \textsuperscript{3} Platform at Ant Group
\vspace{2mm} \\
}

\begin{document}

\maketitle

\begin{abstract}
Training large reasoning models (LRMs) with reinforcement learning in STEM domains is hindered by the scarcity of high‑quality, diverse, and verifiable problem sets. Existing synthesis methods, such as Chain‑of‑Thought prompting, often generate oversimplified or uncheckable data, limiting model advancement on complex tasks. To address these challenges, we introduce \textbf{SHARP}, a unified approach to \textbf{S}ynthesizing \textbf{H}igh‑quality \textbf{A}ligned \textbf{R}easoning \textbf{P}roblems for LRMs reinforcement learning with verifiable rewards (RLVR). \textbf{SHARP} encompasses a strategic set of self‑alignment principles—targeting graduate‑ and Olympiad‑level difficulty, rigorous logical consistency, and unambiguous, verifiable answers—and a structured three‑phase framework (Alignment, Instantiation, Inference) that ensures thematic diversity and fine‑grained control over problem generation. We implement \textbf{SHARP} by leveraging a state‑of‑the‑art LRM to infer and verify challenging STEM questions, then employ a reinforcement learning loop to refine the model’s reasoning through verifiable reward signals. Experiments on benchmarks such as GPQA demonstrate that \textbf{SHARP}‑augmented training substantially outperforms existing methods, markedly improving complex reasoning accuracy and pushing LRM performance closer to expert‑level proficiency. Our contributions include the \textbf{SHARP} strategy, framework design, end‑to‑end implementation, and experimental evaluation of its effectiveness in elevating LRM reasoning capabilities.
\end{abstract}

\section{Introduction}
Large Reasoning Models (LRMs), such as OpenAI-O1, O3/O4 \citep{openai_o1, openai_o3_o4}, Qwen3 \citep{qwen3}, and DeepSeek-R1 \citep{deepseekai2025deepseekr1incentivizingreasoningcapability}, have demonstrated remarkable capabilities in complex domains like mathematics and coding \citep{Chen2025TowardsRE}. However, mastering complex, multi-step reasoning, especially within STEM domains, remains a significant challenge \citep{hochlehnert2025soberlookprogresslanguage, rein2024gpqa, lewkowycz2022solving}. In these fields, models must not only understand the problem but also perform rigorous logical deductions to arrive at accurate answers. While techniques like Chain-of-Thought (CoT) prompting \citep{wei2022chain} encourage models to produce intermediate reasoning steps, the quality, complexity, and logical soundness of these generated paths can be inconsistent, often limited by the scale and quality of the underlying training data. Generating high-quality reasoning data for STEM is notoriously difficult. It requires domain expertise, careful problem construction to avoid ambiguity, and verifiable solutions \citep{lightman2023let}. Manually creating such datasets is expensive and slow, while existing automated methods may lack the necessary depth, diversity, or logical coherence required to train truly advanced reasoning models \citep{deepseekai2025deepseekr1incentivizingreasoningcapability}. This scarcity of suitable training data forms a critical bottleneck in advancing LLM reasoning capabilities towards expert-level or even superintelligence performance \citep{li202512surveyreasoning}.

To overcome these limitations and further enhance LRMs' performance on complex STEM reasoning tasks, particularly those requiring graduate- or Olympiad-level knowledge and reasoning skills (e.g., GPQA), we introduce a novel \textbf{SHARP}(\textbf{S}ynthesizing \textbf{H}igh-quality \textbf{A}ligned \textbf{R}easoning \textbf{P}roblems) approach. Specifically, the main components of \textbf{SHARP} approach include:\\
\textbf{The SHARP Strategy}: The strategy includes a set of self-alignment guiding principles covering problem difficulty (comparable in difficulty to graduate-level coursework or challenging Olympiads), reasoning consistency, answer format, authenticity, language, modality, structure, and output formatting, etc. This strategy focuses not only on alignment principles throughout the reasoning process, but also emphasizes answer verifiability and unambiguity. 

\textbf{The SHARP Framework}: The framework comprise \textbf{Alignment, Instantiation, and Inference} phases. The \textbf{Instantiation} phase includes \textbf{Three-Tier Subject Categorization}, a hierarchical system (e.g., Subject $\to$ Category $\to$ Topic) enabling targeted generation of diverse samples across specific STEM sub-fields. 

\textbf{The SHARP Implementation}: We first implement the \textbf{SHARP} framework leveraging an open-source state-of-the-art LRM (such as DeepSeek R1 \citealp{deepseekai2025deepseekr1incentivizingreasoningcapability}) itself to synthesize self-aligned generative challenging STEM problems systematically, their step-by-step problem-solving reasoning and reference answers, guided by the \textbf{SHARP} strategy. Then we evaluate these samples with general verifiers, such as Math-Verify \citep{huggingface_math-verify}. to obtain the final ground truth. By utilizing these synthesized aligned high-quality and challenging samples, we train large reasoning models through reinforcement learning from zero (like DeepSeek R1 Zero \citep{deepseekai2025deepseekr1incentivizingreasoningcapability}) and further enhance the model's reasoning capabilities in complex STEM problem-solving.

Extensive experiments demonstrate that our proposed \textbf{SHARP} strategy, particularly when coupled with the \textbf{SHARP} framework, through \textbf{SHARP Implementation}, can produce large-scale, high-quality samples capable of significantly boosting the complex reasoning performance of LLMs with reinforcement learning, pushing their reasoning capabilities closer to expert-level proficiency in STEM domains. Our main contributions are as follows:
\begin{itemize}
    \item We propose a novel \textbf{SHARP} approach, comprising a set of carefully designed self-aligned core principles for synthesizing aligned generative complex and high-quality STEM reasoning samples.
    \item We detail the methodology in the \textbf{SHARP} framework, including \textbf{Alignment, Instantiation, and Inference} phases. The \textbf{Instantiation} phase includes a structured data fusion framework incorporating three-tier subject categorization for diverse and targeted sample generation.
    \item We implement the framework for LRMs with reinforcement learning for enhancing the complex reasoning capabilities of STEM problem-solving.
    \item Experiments demonstrate the effectiveness of the proposed approach in improving model performance on challenging STEM reasoning tasks and benchmarks through comprehensive evaluations.
    \item The \textbf{SHARP} offers a potential pathway to significantly enhance LRM performance on challenging STEM reasoning benchmarks like GPQA \citep{rein2024gpqa}.
\end{itemize}

The remainder of this paper is organized as follows: Section 2 introduces background concepts. Section 3 details our proposed \textbf{SHARP} approach. Section 4 outlines the experimental setup. Section 5 presents experimental results and analysis. Section 6 discusses related work. Finally, Section 7 concludes the paper.

\section{Background}
LLMs often struggle with problems demanding true logical reasoning. Optimizing LLM reasoning to enable systematic, human-like logical thinking remains a key research direction. Several techniques are proposed to elicit reasoning from LLMs.

\textbf{Chain-of-Thought (CoT)}: CoT prompting improves LLM performance on complex tasks by guiding them to generate intermediate reasoning steps \citep{wei2022chain, Li2024ChainOT, yeo2025demystifyinglongchainofthoughtreasoning}. By mimicking human thought processes, CoT breaks down complex problems into smaller, manageable steps, aiding comprehension and solution derivation. Variants include Self-Consistency \citep{wang2023selfconsistency}, which samples multiple reasoning paths, and Tree-of-Thoughts \citep{yao2023tree} or Graph-of-Thoughts \citep{Besta_2024}, which explore more complex reasoning structures. However, CoT has limitations: it can be highly dependent on precise prompt engineering. Crucially, the final generated answers cannot easily be verified or even are usually not accurate.
    
\textbf{Self-Alignment in Large Reasoning Models (LRMs)}: Self-alignment utilizes an LLM's own capabilities to refine its behavior or training data \citep{wang2024steponfeet}, aiming to reduce reliance on human annotation and improve data quality and diversity through model self-generation, evaluation, or correction \citep{dong2025selfplay}. Samples include LLMs generating responses to unknown questions with explanations of unanswerability or using multi-round bootstrapping for self-improvement \citep{deng-etal-2024-dont}. Self-alignment offers a promising direction for training more powerful and reliable LLMs \citep{cao2024scalableautomatedalignmentllms}.

\textbf{Reinforcement Learning for LLMs}: The LRMs RL model OpenAI-O1, O3/O4 \citep{openai_o1, openai_o3_o4}, Qwen3 \citep{qwen3}, and DeepSeek-R1 \citep{deepseekai2025deepseekr1incentivizingreasoningcapability} involve self-play or self-critique mechanisms where the model learns from rewards generated based on its own outputs, akin to AlphaZero \citep{silver2017masteringchessshogiselfplay} but applied to text generation and reasoning.

In addition, several challenging benchmark datasets have been developed to evaluate LLM reasoning capabilities in STEM. GPQA (Graduate-Level Google-Proof Q\&A Benchmark) \citep{rein2024gpqa} is designed by domain experts to be extremely difficult (PhDs achieve $\sim$65\% accuracy). Its ``Google-proof'' nature makes it ideal for assessing deep understanding and reasoning, as answers are hard to find via web search. Performance on this benchmark serves as a crucial proxy for evaluating the effectiveness of our proposed \textbf{SHARP} approach.

\section{SHARP: \textbf{\textbf{S}ynthesizing \textbf{H}igh-quality \textbf{A}ligned \textbf{R}easoning \textbf{P}roblems}}
Our proposed \textbf{SHARP} approach aims to systematically generate high-quality, complex STEM reasoning samples by guiding a state-of-the-art LRM (such as DeepSeek R1) instance-alignment reasoning inference through the \textbf{SHARP} framework \ref{sharp_framework} governed by the \textbf{SHARP} following strategy.
\subsection{The \textbf{SHARP} Strategy}
\begin{figure}
    \centering
    \includegraphics[width=1.2\linewidth]{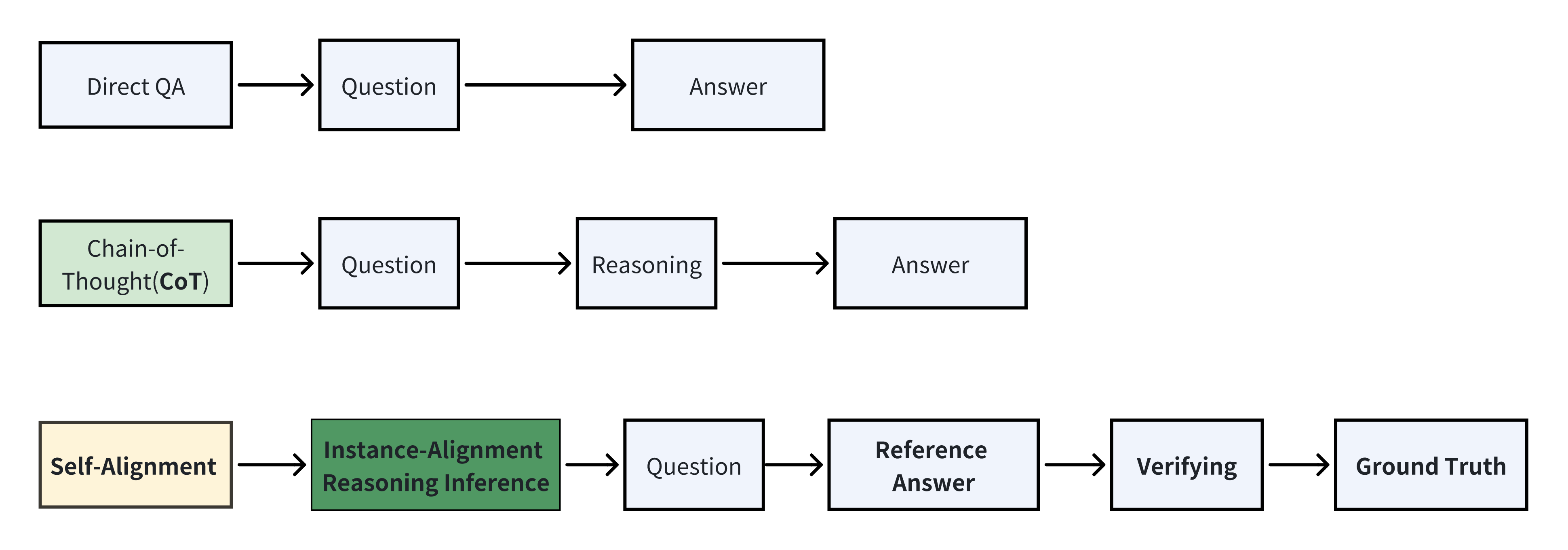}
    \caption{The \textbf{SHARP} Approach}
    \label{fig:sharp_strategy}
\end{figure}

The starting point of the entire \textbf{SHARP} approach is to apply the \textbf{SHARP} strategy, and Fig.\ref{fig:sharp_strategy} illustrates the \textbf{SHARP} strategy pipeline, including instance-level problem generation and alignment inference phases. This indicates that all subsequent steps, especially the \textbf{Instance-Alignment Reasoning Inference} in Fig.\ref{fig:sharp_strategy} (described in \textbf{Instantiation Phase} \ref{sharp_framework}), will strictly follow the self-alignment principles in the \textbf{SHARP} strategy. 

Compared with conventional Direct QA and Chain-of-Thought (CoT) reasoning, the core objective of the \textbf{SHARP} strategy shown in Algo.\ref{alg:SHARP} is to ensure that generated samples possess high-quality and challenging samples, and precise reference answers. These synthesized aligned questions are not only of high difficulty and topic diversity, but also strictly follow the high consistency requirements of logic, ground truth, authenticity, language, structure, modality, and format. More importantly, the verified reference answers of these high-quality questions will strictly meet the Ground Truth consistency and complexity expansion requirements, that is, it will be an objectively verifiable single value (or a specified aggregation form) and follow the format specification.

Specifically, we formalize the \textbf{SHARP} self-alignment strategy as shown in Algorithm \ref{alg:SHARP}.
\begin{algorithm}[t]
\caption{SHARP Self-Alignment Problem Synthesis Strategy}
\label{alg:SHARP}
\begin{algorithmic}[1]
\Require Seed topic set $S = \{s_1, s_2, \ldots, s_N\}$; alignment strategy constraints $\{x_v\}$; base LRM model; reasoning spec $R_{\text{spec}}$; verifier $V$
\Ensure Verified aligned question-answer pairs $Q = \{(q_1, a_1), \ldots, (q_m, a_m)\}$

\State Initialize $Q \gets \emptyset$
\For{each seed topic $s_i \in S$}
    \State \Comment{Alignment Phase}
    \State Configure alignment constraints: Alignment constraints $\{x_v\}$ (See Appendix A for details.)
    \State Construct reasoning blueprint (e.g., step-by-step, propose-verify)
    
    \State \Comment{Instantiation Phase}
    \State Generate prompt $p_i \gets \textproc{InstantiatePrompt}(s_i, \{x_v\}, R_{\text{spec}})$
    \State Select reasoning structure using Three-Tier Category hierarchy.

    \State \Comment{Inference Phase}
    \State Query model with $p_i$ to generate $(q_i, r_i, a_i)$:
    \Statex \hspace{1em} $q_i \gets$ question text, $r_i \gets$ reasoning trace, $a_i \gets$ final answer
 
    \State Format output using SHARP conventions:
    \Statex \hspace{1em} \texttt{<question start>} $q_i$ \texttt{<question end>}
    \Statex \hspace{1em} reasoning: $r_i$, final answer: \textbackslash boxed\{\{\$answer\}\}

    \State \Comment{Verifying Phase}
    \If{$V(r_i, a_i)$ passes all alignment checks}
        \State $Q \gets Q \cup \{(q_i, a_i)\}$
    \EndIf
\EndFor
\State \Return $Q$
\end{algorithmic}
\end{algorithm}

\subsection{The \textbf{SHARP} Framework}{\label{sharp_framework}}
Building upon the \textbf{SHARP} strategy, we introduce an enhanced \textbf{SHARP} data fusion framework specifically designed for synthesizing high-quality reasoning problems in STEM sub-disciplines. The core of this framework is the construction of the ``\textbf{Seed Topics library}'', which is built on a ``Three-Tier Category'' knowledge structure. This structure integrates the Magpie query generation approach \citep{xu2025magpie} with advanced semantic clustering and balanced sampling techniques, improving both the diversity and representativeness of the synthetic reasoning queries. Seed documents are meticulously curated from established benchmark question banks (we will not directly rephrase the query based on the validation set, but only analyze the topic keypoints covered by these benchmarks) and high-quality handcrafted corpora (STEM textbooks, papers, and data recalled through Common Crawl etc.), while cutting-edge LLMs, such as DeepSeek R1 and Qwen3, are employed to facilitate comprehensive topic extraction and ``Three-Tier Category'' generation, ensuring a broad coverage of critical reasoning domains. The clustering process, utilizing K-means algorithms \citep{macqueen1967some} on BGE-m3 embeddings \citep{chen2024bgem3embeddingmultilingualmultifunctionality}, in tandem with balanced sampling, addresses potential biases, ensuring uniform representation across a spectrum of reasoning topics.

Moreover, the integration of persona-based methodologies \citep{ge2025scalingsyntheticdatacreation} and keypoint enhancements introduces a diverse array of reasoning contexts and enables the modulation of query difficulty levels, facilitating the generation of training data that reflects both problem complexity and cognitive challenges. This methodological approach ensures scalable reasoning problem synthesis that aligns closely with the depth and complexity required in STEM-related tasks. The \textbf{SHARP} framework, by leveraging sophisticated reasoning capabilities of LLMs like DeepSeek R1, synthesizes logically coherent, complex reasoning problems that are carefully aligned with the nuanced demands of STEM disciplines. The primary objective of this framework is to generate high-quality, diverse training samples that drive the optimization of reinforcement learning (RL) models, especially in the context of high-difficulty STEM benchmarks.

\begin{figure}
    \centering
    \includegraphics[width=1.2\linewidth]{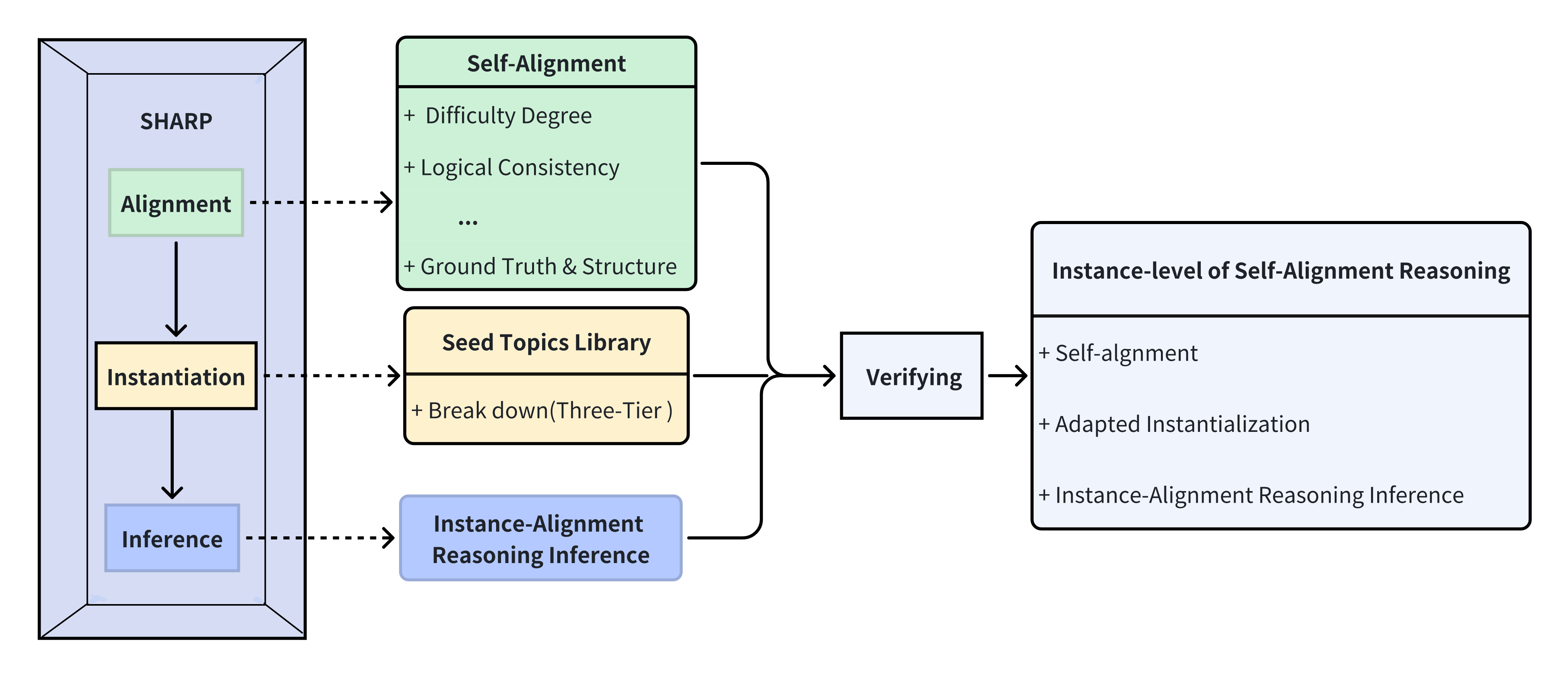}
    \caption{The \textbf{SHARP} Framework}
    \label{fig:sharp_framework}
\end{figure}

The \textbf{SHARP} framework is underpinned by its core modules, prominently featuring the \textbf{Self-Alignment} block (depicted in Fig. \ref{fig:sharp_framework}). This block guarantees the adherence to stringent quality standards for the generated content, encompassing aspects such as question difficulty, reasoning consistency, and answer verifiability. It systematically encompasses three fundamental phases: \textbf{Alignment}, \textbf{Instantiation}, and \textbf{Implement}, thereby constructing a comprehensive reasoning alignment pathway from the initial \textbf{SHARP} strategy formulation to the generation of specific instances. This structured approach ensures that all internal operations are cohesively aligned with the \textbf{SHARP} strategy, ultimately facilitating the synthesis of high-quality, aligned reasoning problems tailored for reinforcement learning in large-scale reasoning models. The three phases of the \textbf{SHARP} framework are detailed as follows.

\textbf{Alignment Phase}: This phase initiates the \textbf{SHARP} approach, and serves as its implicit input of the overall goal of applying the \textbf{SHARP} Algo.\ref{alg:SHARP} strategy, corresponding to the ``\textbf{Self-Alignment}'' detail box on the right. The specific requirements set in this phase are the key manifestation of the \textbf{SHARP} Algo.\ref{alg:SHARP} strategy in sample generation, and all subsequent steps will align strictly to it. It inherits and strengthens the core advantages of \textbf{Self-Alignment}, especially the structural requirements for the reasoning process, ensuring the reasoning consistency and reliability of the generated samples, and helping the training model to form more standardized and reliable reasoning capabilities.

We begin this phase by operationalizing the \textbf{SHARP} principles into executable constraints, passing specific requirements (e.g., difficulty level, reasoning style, verification method, etc.) to the next phase. Then the \textbf{SHARP} plans a systematic reasoning framework or blueprint that meets logical consistency, ensuring that each step of deduction is supported by STEM theory or logic, eliminating jumps and intuitive guesses, and maintaining format requirements (such as the Math-Verify \citep{huggingface_math-verify}). It enforces that reasoning must be planned, orderly, and verifiable, rather than arbitrary heuristic deduction. For example, we set the ``Difficulty Degree'' to graduate- or Olympiad-level, mandate a ``Step-by-Step'' reasoning process, and employ a ``Propose-verify'' mechanism where the model internally proposes and verifies each reasoning step for validity and truthfulness. These standards are consistent with those in the \textbf{Verifying} stage. 

\textbf{Instantiation Phase}: Building on the \textbf{SHARP} Algo.\ref{alg:SHARP} strategy of the \textbf{Alignment Phase} and a ``Three-Tier Category'' knowledge framework, a clear ``reasoning structure'' definition stage for the instantiations is introduced, distinct from the relatively free ``Reasoning'' step in traditional CoT. This instantiation phase integrates a ``Three-Tier Category'' knowledge framework to instantiate the strategy to different subjects' characteristics and structures. The ``Three-Tier Category'' knowledge framework manages and organizes STEM knowledge hierarchically (e.g., Chemistry $\to$ Organic Chemistry, Spectroscopy $\to$ Elimination reactions, IR spectroscopy, Carbonyl compounds, Alcohols, Characteristic IR absorption frequencies). A comprehensive and easily expandable ``\textbf{Seed Topics library}'' (orange detail box) is organized in this hierarchy. This ensures the combination of broad thematic coverage and professional depth, enabling targeted generation of complex samples in specific STEM sub-fields. The structured topic information informs the \textbf{Verifying} stage for confirming instance topic attribution and generation relevance.

\textbf{Inference Phase}: Under the guidance of the ``reasoning structure'' defined in the \textbf{Instantiation Phase}, an instantiated aligned reasoning inference process generating a specific STEM sample that meets the \textbf{SHARP} strategy is performed. This involves leveraging the capabilities of the state-of-the-art LRMs model (such as DeepSeek R1) to generate aligned reasoning instances. Then these aligned reasoning instances are submitted to the next \textbf{Verifying} stage.

The \textbf{Verifying} stage is mainly responsible for quality control, strictly verifying whether the adapted instance fully complies with the \textbf{Self-Alignment} detail box, and whether it is consistent with the three-level category theme of \textbf{All Seed Topics} it claims, to ensure that the final output sample meets the preset high standards detailed in \textbf{SHARP} strategy. The final output of the entire process - high-quality samples are generated that can elicit complex reasoning for LRMs RL training. 

Building upon \textbf{SHARP} Algo.\ref{alg:SHARP}, \textbf{SHARP} framework \ref{fig:sharp_framework} introduces innovative ``instance-level'' reasoning, where each sample constitutes a complete and self-consistent reasoning structure. This is achieved through a refined three-level subject classification adaptation mechanism, a robust inference and verification process. These meticulously refined samples are invaluable for the model, enabling it to learn fine-grained knowledge and complex reasoning patterns. By systematically generating STEM samples with ultra-high complexity at the sample level, this comprehensive approach provides unique value and significant potential for enhancing the complex reasoning capabilities of LRMs, particularly in improving top STEM reasoning benchmarks, such as GPQA.

\subsection{The \textbf{SHARP} Implementation}
\begin{figure}
    \centering
    \includegraphics[width=1\linewidth]{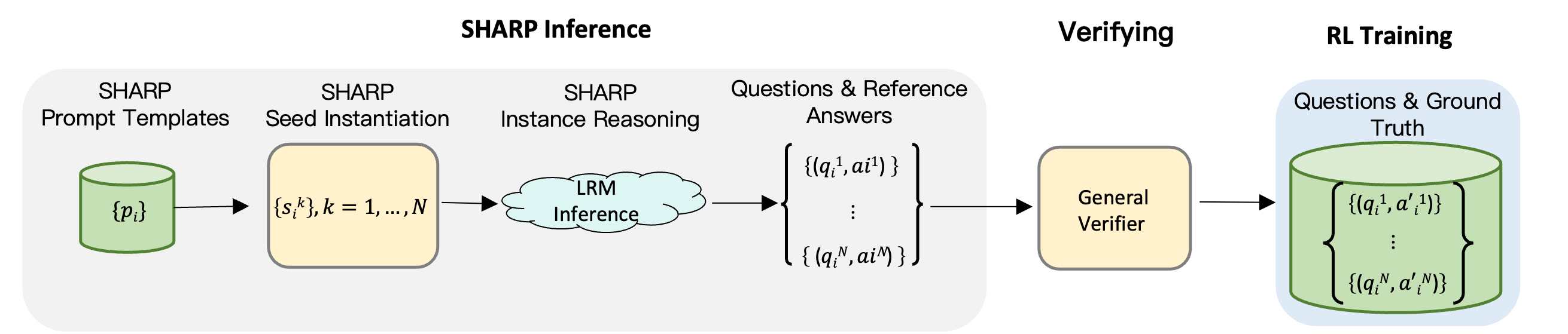}
    \caption{The \textbf{SHARP} Implementation for Large Reasoning Models Reinforcement Learning}
    \label{fig:sharp_reasoning}
\end{figure}

Based on the \textbf{SHARP} strategy and framework, we implement the \textbf{SHARP} approach with a state-of-the-art LRM model (like DeepSeek R1) and generate complex reasoning samples at the sample level, shown in Fig. \ref{fig:sharp_framework}. Then these synthesized high-quality self-aligned samples are used to enhance the complex reasoning ability of the RL Zero model training (inspired by RL Zero \citep{openai_o1, deepseekai2025deepseekr1incentivizingreasoningcapability, openai_o3_o4}). The system mainly consists of three core stages:\textbf{ SHARP Inference}, \textbf{Verifying} and \textbf{RL Zero Training} and are detailed as follows.

\textbf{SHARP Inference}:
\begin{enumerate}
\item \textbf{SHARP Prompt Templates ($\{p_i\}$}): As the starting point of the process, initial aligned question prompt templates that meet the \textbf{SHARP} strategy (such as high difficulty, plain text, single question, factual accuracy, etc.) are constructed.

\item \textbf{SHARP Seed Instantiation ($\{s_i^k\},  k=1,..., N$}, here $N$ is the total number of topics): Extract an aligned prompt $x_i$ from the \textbf{SHARP} prompt templates, and map the input general prompt $x_i$ with the specific STEM knowledge points or topic ``seeds'' $\{s_i^k\}$ based on the ``\textbf{Seed Topics Library}'' according to  ``Three-Tier Category'' knowledge framework in the \textbf{SHARP} framework. This provides context for the subsequent generation of domain-specific and depth-specific reasoning, and the output is the prompt $x_i$ and its associated instance with adapted specific topic $\{s_i^k\}$ from \textbf{Seed Topics Library} in the ``\textbf{Instantiation Phase}''.

\item \textbf{Questions and Reference Answers $(\{(q_i^k, a_i^k), k=1,..., m\})$}: Use the current best version of the reasoning model (e.g., DeepSeek R1) to generate candidate reasoning responses for the prompts and topic contexts, including high-quality questions $\{q_i^k\}$, its \textbf{SHARP} reasoning processes, and corresponding candidate reference answers $(\{a_i^k\})$).
\end{enumerate}

\textbf{Verifying}: Verification is performed according to the \textbf{SHARP} strategy and a scoring rubric combining model self-check confidence and a rule-based reward model for the generated question and answer pairs $(\{(q_i^k, a_i^k)\})$. The purpose is to further screen out data with poor quality, logical errors, unreliable reward signals, or data that does not meet the final requirements, and ensure that only the highest quality and most reliable samples enter the LRM RL training stage.

\textbf{RL Training}: Using the high-quality and verified samples created by the \textbf{SHARP Inference} and \textbf{Verifying}, RL Zero training is carried out based on an RL algorithm (e.g., PPO \citep{schulman2017proximalpolicyoptimizationalgorithms}, or GRPO \citep{shao2024deepseekmathpushinglimitsmathematical}).

The unique value of \textbf{SHARP} inference system lies in its focus on quality and complexity, adaptive sample generation of higher quality and more complex reasoning samples, and thereby can train a more powerful LRM model through RL Zero and push the upper limit of the model's reasoning capabilities in STEM fields dimensions.

\section{Experimental Setup}
Our experimental setup was meticulously designed to rigorously evaluate the \textbf{SHARP} approach. The Training Data consisted of two primary sets: a baseline dataset generated using standard CoT prompting on existing STEM samples and a substantial dataset of 190,000 samples generated via the \textbf{SHARP} methodology. Our comparison models included distinct sets for distillation training and RL Zero training. For distill training, the \textbf{Qwen2.5-7B-Instruct-Distill} model served as a baseline, representing capable LRMs without specific STEM reasoning dataset training. This was compared against state-of-the-art \textbf{DeepSeek-R1-Distill-Qwen-7B} \citep{deepseekai2025deepseekr1incentivizingreasoningcapability} and \textbf{SHARP-Qwen2.5-7B-Instruct-Distill} \citep{qwen2025qwen25technicalreport}, where the latter was the baseline model further distilled on \textbf{SHARP}-generated and verified samples. In the RL Zero Training comparison, \textbf{Open-Reasoner-Zero-7B} \citep{hu2025openreasonerzeroopensourceapproach} was the baseline, evaluated against \textbf{SHARP-Open-Reasoner-Zero-7B}, which was trained using \textbf{SHARP}-generated problems through an RL Zero process. The training details for distillation involved standard procedures on the respective datasets. For \textbf{SHARP}-RL Zero training, we employed the GRPO algorithm, with a rule-based reward function, alongside specified hyperparameters and computational resources detailed in the appendix. Finally, evaluation metrics centered on model performance on the challenging GPQA STEM reasoning benchmark, using accuracy metrics like pass@k to compare models trained with and without \textbf{SHARP}-generated samples, thereby demonstrating the efficacy of our approach.

\section{Experiments: Results and Analysis}
Building upon the described experimental setup, our evaluations demonstrate the significant advantages of the \textbf{SHARP} methodology in enhancing large reasoning models. The experiments were conducted in two primary modes: high-quality complex reasoning knowledge distillation with supervised fine-tuning, and the utilization of challenging \textbf{SHARP}-generated samples to elicit complex reasoning capabilities in LRMs.

The results, presented in Table \ref{table:performance_comparison} and \ref{table:evaluation_7b}, are compelling. Approximately 190,000 STEM samples were constructed using the \textbf{SHARP} approach. In the distillation experiments (Table \ref{table:performance_comparison}), the \textbf{SHARP-Qwen2.5-7B-Instruct-Distill} model, trained on \textbf{SHARP} data, achieved a GPQA Diamond score of 54.7. This represents an 8.3 percentage point improvement over the \textbf{Qwen2.5-7B-Instruct-Distill} baseline (46.4) and a 4.8 percentage point increase over the \textbf{DeepSeek-R1-Distill-Qwen-7B model} (49.9). This notable outperformance, even without RL refinement, underscores the superior quality of data generated by the structured \textbf{SHARP} approach. The \textbf{SHARP}-trained model also showed consistent improvements across GPQA Physics (71.1 vs. 60.6 baseline), Chemistry (38.8 vs. 31.3 baseline), and Biology (57.9 vs. 55.9 baseline).

\begin{table}[h!]
\centering
\resizebox{\textwidth}{!}{
\begin{tabular}{lcccc}
\toprule
Models & GPQA Physics & GPQA Chemistry & GPQA Biology & GPQA Diamond \\
\midrule
Qwen2.5-7B-Instruct-Distill (Baseline) & 60.6 & 31.3 & 55.9 & 46.4 \\
DeepSeek-R1-Distill-Qwen-7B & 70.1 & 31.9 & 43.4 & 49.9 \\
\midrule
\rowcolor{green!20}
\textbf{SHARP}-Qwen2.5-7B-Instruct-Distill & 71.1 & 38.8 & 57.9 & 54.7 \\
\bottomrule
\end{tabular}}
\caption{Performance on GPQA benchmark (Diamond subset: most difficult tier), comparing distilled models trained with and without \textbf{SHARP}-synthesized data.}
\label{table:performance_comparison}
\end{table}

\begin{table}[h!]
\centering
\resizebox{\textwidth}{!}{
\begin{tabular}{lcccc}
\toprule
Models & GPQA Physics & GPQA Chemistry & GPQA Biology & GPQA Diamond \\
\midrule
Open-Reasoner-Zero-7B (Baseline) & 41.4 & 27.4 & 48.7 & 35.5 \\
\midrule
\rowcolor{green!20}
\textbf{SHARP}-Open-Reasoner-Zero-7B & 44.6 & 26.3 & 54.9 & 37.0 \\
\bottomrule
\end{tabular}}
\caption{Performance on GPQA benchmark (Diamond subset: most difficult tier), comparing RL Zero models trained with and without \textbf{SHARP}-synthesized data.}
\label{table:evaluation_7b}
\end{table}

In the RL-Zero reasoning training experiments (Table \ref{table:evaluation_7b}), the \textbf{SHARP-Open-Reasoner-Zero-7B} model, leveraging SHARP-generated STEM problems, achieved a GPQA Diamond score of 37.0, marking a 1.5 percentage point improvement over the \textbf{Open-Reasoner-Zero-7B} baseline (35.5). This outcome offers initial validation for the efficacy of \textbf{SHARP}-synthesized data in supporting RL-Zero reasoning training. Notably, performance enhancements were recorded in GPQA Physics (44.6 vs. 41.4 baseline) and GPQA Biology (54.9 vs. 48.7 baseline). Conversely, GPQA Chemistry exhibited a marginal decrease (26.3 vs. 27.4 baseline). We attribute this to the inherently high dependence of chemistry problems on deep, structured domain knowledge and nuanced symbolic reasoning, which may not be as effectively acquired through unsupervised RL Zero methods without pre-distilled domain-specific priors, as detailed in Appendix C.2. Detailed analyses of these discrepancies, including sample difficulty metrics (e.g., response length and reward signal distribution), are provided in Appendix B.2.

To further substantiate these findings and provide a more granular understanding, Appendix B includes comparative evaluations of \textbf{SHARP}-based distillation and ablation studies across STEM fields and mathematical data. Specifically, we present controlled experiments analyzing the performance impact of different \textbf{SHARP}-generated sub-corpora—physics, chemistry, biology—on both distillation and RL Zero models. Representative subject-level ablation examples further demonstrate how \textbf{SHARP}’s three-tier taxonomy enables precise control over difficulty and topic diversity. Additionally, Appendix C systematically evaluates the 190,000 \textbf{SHARP}-generated samples, showcasing balanced distributions across 600+ granular STEM subcategories and pass rate analyses that correlate with human expert assessments. Together, these results confirm that \textbf{SHARP}’s aligned and structured synthesis framework successfully generates high-difficulty, verifiable problems—spanning quantum mechanics to organic reaction mechanisms—that directly enhance LRMs’ capacity for expert-level scientific reasoning.

Collectively, these findings highlight the \textbf{SHARP} approach's effectiveness in generating high-quality, complex training samples. The consistent performance gains observed across different models and evaluation subjects, particularly on the demanding GPQA-Diamond set, demonstrate that \textbf{SHARP} significantly enhances the capability of LRMs to tackle complex STEM reasoning tasks, pushing their performance closer to expert-level proficiency. The structured generation process, guided by \textbf{SHARP}'s self-alignment strategy, yields problems that is not only diverse and challenging but also logically rigorous and verifiable, directly contributing to the observed improvements in LLMs' reasoning abilities.

\section{Related Work}
\textbf{LLM Reasoning Enhancement}: As discussed in Section 2, numerous efforts focus on improving LLM reasoning via prompting (CoT, ToT, GoT) \citep{wei2022chain, yao2023tree, Besta_2024} or specialized fine-tuning \citep{trung-etal-2024-reft, lobo-etal-2025-impact}. However, scaling up verifiable signals for long CoT remains challenging due to the limited availability of high-quality, verifiable samples \citep{yeo2025demystifyinglongchainofthoughtreasoning}. In contrast, \textbf{SHARP} generates verifiable, high-quality samples without relying on prompt engineering, enabled by structured self-alignment.

\textbf{Synthetic Data for LLM Reasoning}: Synthesizing data plays a crucial role in training large language models (LLMs) to enhance their reasoning abilities. Approaches like Self-Instruct and Alpaca \citep{wang2023selfinstructaligninglanguagemodels} have pioneered the use of generated instructional data to align LLM behaviors with desired outcomes. \citep{Shao2023SyntheticPG} introduced a method where a limited set of handcrafted samples prompts the model to autonomously create additional data, selectively incorporating high-quality demonstrations to bolster reasoning performance. Nemotron-CrossThink \citep{akter2025nemotroncrossthinkscalingselflearningmath} leverages cross-thought reasoning to enable self-improvement within mathematical domains, while Qwen2.5-Math and Qwen2.5-Coder \citep{yang2024qwen25mathtechnicalreportmathematical, hui2024qwen25codertechnicalreport} focus on generating domain-specific data for mathematical problem-solving and coding tasks, respectively. Phi-4-Reasoning \citep{xu2025phi4minireasoningexploringlimitssmall, abdin2025phi4reasoningtechnicalreport} demonstrates the effectiveness of compact architectures in handling complex reasoning tasks. \citep{goldie2025syntheticdatageneration} introduced SWiRL for multi-step reward shaping; such signal shaping is partially mirrored in our SHARP RL reward design. Together, these studies highlight the significance of sophisticated data synthesis strategies in improving LLM reasoning capabilities. Unlike these approaches, \textbf{SHARP} specifically targets the synthesis of challenging STEM problems by enforcing a unique combination of explicit self-alignment principles for reasoning consistency, thematic diversity, and strict answer verifiability, aiming to overcome the limitations in generating consistently complex and reliable reasoning samples.

\textbf{Self-Alignment.} Self-alignment refers to training paradigms that utilize a model’s own capabilities to assess, revise, or supervise its outputs—reducing the reliance on external human annotation. Existing approaches can be broadly categorized into three paradigms. First, \emph{preference-based self-alignment} leverages internal comparisons or preference signals, often sparse or noisy, to guide alignment. For example, the Hummer framework \citep{wu2024hummer} investigates competitive learning under weak preference supervision, emphasizing the importance of carefully designed reward structures for stable alignment. Second, \emph{consistency-based self-alignment} uses internal logical coherence as a proxy for correctness. The SelfFeedback framework \citep{liang2024internalconsistencyselffeedbacklarge} enhances model reasoning by identifying and reinforcing internally consistent outputs, offering a form of self-improvement without external labels. Third, and most relevant to our work, are \emph{verifiability-based self-alignment} approaches, which aim to provide strong and objective training signals by enforcing explicit correctness. Our proposed \textbf{SHARP} framework builds on the motivation of reducing human oversight but introduces a structured, multi-phase self-alignment strategy tailored to complex STEM domains. Unlike methods that rely on implicit or heuristic feedback, \textbf{SHARP} synthesizes \emph{explicitly verifiable}, \emph{logically rigorous}, and \emph{thematically diverse} reasoning samples that are suitable for use as direct supervision targets. By encoding these properties through a principled three-phase pipeline—\textbf{Alignment, Instantiation, and Inference}—\textbf{SHARP} enables the systematic synthesis of reasoning problems that are not only logically rigorous and thematically diverse, but also explicitly verifiable. This structure supports high-fidelity, reward-aligned supervision signals, making it particularly effective for reinforcement learning from verifiable rewards (RLVR) in domains where correctness must be grounded in domain knowledge and formal reasoning. In this way, \textbf{SHARP} bridges the methodological gap between implicit alignment strategies and the stringent verification demands of complex STEM problem-solving.

In summary, \textbf{SHARP} offers a principled and extensible framework that advances the current state of self-alignment and synthetic reasoning for large language models. By tightly integrating explicit alignment objectives with a structured, verifiability-driven sample generation process, \textbf{SHARP} produces high-difficulty and semantically controlled reasoning data that overcomes the limitations of heuristic prompting, weak preference modeling, or unverified self-consistency. This makes \textbf{SHARP} especially well-suited for reinforcement learning settings in scientific domains, where both logical fidelity and verifiable correctness are essential. Ultimately, \textbf{SHARP} contributes a scalable, high-precision foundation for training advanced reasoning models to operate effectively in complex and rigorous STEM contexts.

\section{Conclusion, Limitations and Future Work}
We presented a novel \textbf{SHARP} approach to address the critical need for high-quality, complex, and verifiable training problems for enhancing the reasoning capabilities of LLMs, particularly in STEM domains. By employing \textbf{SHARP} inference and \textbf{Verifying} process, our approach systematically guides LRMs to generate challenging problems and logically sound, verifiable solutions efficiently and at scale, addressing the limitations of traditional CoT methods in producing difficult, diverse, and logically rigorous STEM reasoning samples. We presented the \textbf{SHARP} inference integrating with \textbf{Verifying} process, enabling iterative RL foundation model training and performance enhancement on complex reasoning tasks. Experimental results demonstrate significant performance gains on challenging STEM benchmark GPQA compared to baselines trained on CoT data and public STEM datasets, as well as substantial improvement over the state-of-the-art baseline model. For instance, SHARP-augmented distillation training resulted in an 8.3 percentage point improvement on the GPQA Diamond benchmark over the baseline. This validates the effectiveness of our proposed approach in enhancing the ability of large reasoning models to tackle complex STEM problems. 

Future work could explore applying this approach to other domains and more complex reasoning tasks, and further optimizing the \textbf{SHARP} approach on various larger-scale RL reasoning foundation models. Besides, designing a reward function that weights principles from the \textbf{SHARP} strategy will be carried out. And distinctions among different subjects, such as chemistry and biology, have different subject attributes from physics and mathematics, which may involve the further improvement of logic, knowledge graph, and symbolic reasoning capabilities.

\bibliography{sharp}

\begin{thebibliography}{53}
\providecommand{\natexlab}[1]{#1}
\providecommand{\url}[1]{\texttt{#1}}
\expandafter\ifx\csname urlstyle\endcsname\relax
  \providecommand{\doi}[1]{doi: #1}\else
  \providecommand{\doi}{doi: \begingroup \urlstyle{rm}\Url}\fi

\bibitem[Abdin et~al.(2025)Abdin, Agarwal, Awadallah, Balachandran, Behl, Chen,
  de~Rosa, Gunasekar, Javaheripi, Joshi, Kauffmann, Lara, Mendes, Mitra, Nushi,
  Papailiopoulos, Saarikivi, Shah, Shrivastava, Vineet, Wu, Yousefi, and
  Zheng]{abdin2025phi4reasoningtechnicalreport}
Marah Abdin, Sahaj Agarwal, Ahmed Awadallah, Vidhisha Balachandran, Harkirat
  Behl, Lingjiao Chen, Gustavo de~Rosa, Suriya Gunasekar, Mojan Javaheripi,
  Neel Joshi, Piero Kauffmann, Yash Lara, Caio César~Teodoro Mendes, Arindam
  Mitra, Besmira Nushi, Dimitris Papailiopoulos, Olli Saarikivi, Shital Shah,
  Vaishnavi Shrivastava, Vibhav Vineet, Yue Wu, Safoora Yousefi, and Guoqing
  Zheng.
\newblock Phi-4-reasoning technical report, 2025.
\newblock URL \url{https://arxiv.org/abs/2504.21318}.

\bibitem[Akter et~al.(2025)Akter, Prabhumoye, Novikov, Han, Lin, Bakhturina,
  Nyberg, Choi, Patwary, Shoeybi, and
  Catanzaro]{akter2025nemotroncrossthinkscalingselflearningmath}
Syeda~Nahida Akter, Shrimai Prabhumoye, Matvei Novikov, Seungju Han, Ying Lin,
  Evelina Bakhturina, Eric Nyberg, Yejin Choi, Mostofa Patwary, Mohammad
  Shoeybi, and Bryan Catanzaro.
\newblock Nemotron-crossthink: Scaling self-learning beyond math reasoning,
  2025.
\newblock URL \url{https://arxiv.org/abs/2504.13941}.

\bibitem[ArtofProblemSolving()]{artofproblemsolving}
ArtofProblemSolving.
\newblock Art of problem solving community.
\newblock \url{https://artofproblemsolving.com/community/}.

\bibitem[Besta et~al.(2024)Besta, Blach, Kubicek, Gerstenberger, Podstawski,
  Gianinazzi, Gajda, Lehmann, Niewiadomski, Nyczyk, and Hoefler]{Besta_2024}
Maciej Besta, Nils Blach, Ales Kubicek, Robert Gerstenberger, Michal
  Podstawski, Lukas Gianinazzi, Joanna Gajda, Tomasz Lehmann, Hubert
  Niewiadomski, Piotr Nyczyk, and Torsten Hoefler.
\newblock Graph of thoughts: Solving elaborate problems with large language
  models.
\newblock \emph{Proceedings of the AAAI Conference on Artificial Intelligence},
  38\penalty0 (16):\penalty0 17682–17690, March 2024.
\newblock ISSN 2159-5399.
\newblock \doi{10.1609/aaai.v38i16.29720}.
\newblock URL \url{http://dx.doi.org/10.1609/aaai.v38i16.29720}.

\bibitem[{Book Industry Study Group}(2025)]{bisg_bisac}
{Book Industry Study Group}.
\newblock {BISAC Subject Headings}, 2025.
\newblock URL \url{https://www.bisg.org/complete-bisac-subject-headings-list}.

\bibitem[Cao et~al.(2024)Cao, Lu, Lu, Chen, Ren, Xiang, Liu, Lu, He, Han, Sun,
  Lin, and Yu]{cao2024scalableautomatedalignmentllms}
Boxi Cao, Keming Lu, Xinyu Lu, Jiawei Chen, Mengjie Ren, Hao Xiang, Peilin Liu,
  Yaojie Lu, Ben He, Xianpei Han, Le~Sun, Hongyu Lin, and Bowen Yu.
\newblock Towards scalable automated alignment of llms: A survey, 2024.
\newblock URL \url{https://arxiv.org/abs/2406.01252}.

\bibitem[Chen et~al.(2024)Chen, Xiao, Zhang, Luo, Lian, and
  Liu]{chen2024bgem3embeddingmultilingualmultifunctionality}
Jianlyu Chen, Shitao Xiao, Peitian Zhang, Kun Luo, Defu Lian, and Zheng Liu.
\newblock {M}3-embedding: Multi-linguality, multi-functionality,
  multi-granularity text embeddings through self-knowledge distillation.
\newblock In Lun-Wei Ku, Andre Martins, and Vivek Srikumar (eds.),
  \emph{Findings of the Association for Computational Linguistics: ACL 2024},
  pp.\  2318--2335, Bangkok, Thailand, August 2024. Association for
  Computational Linguistics.
\newblock \doi{10.18653/v1/2024.findings-acl.137}.
\newblock URL \url{https://aclanthology.org/2024.findings-acl.137/}.

\bibitem[Chen et~al.(2025)Chen, Qin, Liu, Peng, Guan, Wang, Hu, Zhou, Gao, and
  Che]{Chen2025TowardsRE}
Qiguang Chen, Libo Qin, Jinhao Liu, Dengyun Peng, Jiannan Guan, Peng Wang,
  Mengkang Hu, Yuhang Zhou, Te~Gao, and Wanxiang Che.
\newblock Towards reasoning era: A survey of long chain-of-thought for
  reasoning large language models.
\newblock \emph{arXiv preprint arXiv:2503.09567}, 2025.

\bibitem[DeepScaleR(2025)]{deepscaler}
DeepScaleR.
\newblock Deepscaler-preview-dataset, 2025.
\newblock URL
  \url{https://huggingface.co/datasets/math-dataset/DeepScaleR-Preview-Dataset}.

\bibitem[DeepSeek-AI(2025)]{deepseekai2025deepseekr1incentivizingreasoningcapability}
DeepSeek-AI.
\newblock Deepseek-r1: Incentivizing reasoning capability in llms via
  reinforcement learning, 2025.
\newblock URL \url{https://arxiv.org/abs/2501.12948}.

\bibitem[Deng et~al.(2024)Deng, Zhao, Li, Ng, and Chua]{deng-etal-2024-dont}
Yang Deng, Yong Zhao, Moxin Li, See-Kiong Ng, and Tat-Seng Chua.
\newblock Don`t just say {\textquotedblleft}{I} don`t know{\textquotedblright}!
  self-aligning large language models for responding to unknown questions with
  explanations.
\newblock In Yaser Al-Onaizan, Mohit Bansal, and Yun-Nung Chen (eds.),
  \emph{Proceedings of the 2024 Conference on Empirical Methods in Natural
  Language Processing}, pp.\  13652--13673, Miami, Florida, USA, November 2024.
  Association for Computational Linguistics.
\newblock \doi{10.18653/v1/2024.emnlp-main.757}.
\newblock URL \url{https://aclanthology.org/2024.emnlp-main.757/}.

\bibitem[Dong et~al.(2025)Dong, Lu, Li, Xia, Yu, Zhou, and
  Zhou]{dong2025selfplay}
Guanting Dong, Keming Lu, Chengpeng Li, Tingyu Xia, Bowen Yu, Chang Zhou, and
  Jingren Zhou.
\newblock Self-play with execution feedback: Improving instruction-following
  capabilities of large language models.
\newblock In \emph{The Thirteenth International Conference on Learning
  Representations}, 2025.
\newblock URL \url{https://openreview.net/forum?id=cRR0oDFEBC}.

\bibitem[Ge et~al.(2025)Ge, Chan, Wang, Yu, Mi, and
  Yu]{ge2025scalingsyntheticdatacreation}
Tao Ge, Xin Chan, Xiaoyang Wang, Dian Yu, Haitao Mi, and Dong Yu.
\newblock Scaling synthetic data creation with 1,000,000,000 personas, 2025.
\newblock URL \url{https://arxiv.org/abs/2406.20094}.

\bibitem[Goldie et~al.(2025)Goldie, Mirhoseini, Zhou, Cai, and
  Manning]{goldie2025syntheticdatageneration}
Anna Goldie, Azalia Mirhoseini, Hao Zhou, Irene Cai, and Christopher~D.
  Manning.
\newblock Synthetic data generation \& multi-step rl for reasoning \& tool use,
  2025.
\newblock URL \url{https://arxiv.org/abs/2504.04736}.

\bibitem[Hochlehnert et~al.(2025)Hochlehnert, Bhatnagar, Udandarao, Albanie,
  Prabhu, and Bethge]{hochlehnert2025soberlookprogresslanguage}
Andreas Hochlehnert, Hardik Bhatnagar, Vishaal Udandarao, Samuel Albanie, Ameya
  Prabhu, and Matthias Bethge.
\newblock A sober look at progress in language model reasoning: Pitfalls and
  paths to reproducibility, 2025.
\newblock URL \url{https://arxiv.org/abs/2504.07086}.

\bibitem[Hu et~al.(2024)Hu, Wu, Zhu, Xianyu, Wang, Zhang, and
  Cao]{hu2024openrlhfeasytousescalablehighperformance}
Jian Hu, Xibin Wu, Zilin Zhu, Xianyu, Weixun Wang, Dehao Zhang, and Yu~Cao.
\newblock Openrlhf: An easy-to-use, scalable and high-performance rlhf
  framework, 2024.
\newblock URL \url{https://arxiv.org/abs/2405.11143}.

\bibitem[Hu et~al.(2025)Hu, Zhang, Han, Jiang, Zhang, and
  Shum]{hu2025openreasonerzeroopensourceapproach}
Jingcheng Hu, Yinmin Zhang, Qi~Han, Daxin Jiang, Xiangyu Zhang, and Heung-Yeung
  Shum.
\newblock Open-reasoner-zero: An open source approach to scaling up
  reinforcement learning on the base model, 2025.
\newblock URL \url{https://arxiv.org/abs/2503.24290}.

\bibitem[HuggingFace()]{huggingface_math-verify}
HuggingFace.
\newblock Math-verify.
\newblock \url{https://github.com/huggingface/Math-Verify}.

\bibitem[Hui et~al.(2024)Hui, Yang, Cui, Yang, Liu, Zhang, Liu, Zhang, Yu, Lu,
  Dang, Fan, Zhang, Yang, Men, Huang, Zheng, Miao, Quan, Feng, Ren, Ren, Zhou,
  and Lin]{hui2024qwen25codertechnicalreport}
Binyuan Hui, Jian Yang, Zeyu Cui, Jiaxi Yang, Dayiheng Liu, Lei Zhang, Tianyu
  Liu, Jiajun Zhang, Bowen Yu, Keming Lu, Kai Dang, Yang Fan, Yichang Zhang,
  An~Yang, Rui Men, Fei Huang, Bo~Zheng, Yibo Miao, Shanghaoran Quan, Yunlong
  Feng, Xingzhang Ren, Xuancheng Ren, Jingren Zhou, and Junyang Lin.
\newblock Qwen2.5-coder technical report, 2024.
\newblock URL \url{https://arxiv.org/abs/2409.12186}.

\bibitem[Kwon et~al.(2023)Kwon, Li, Zhuang, Sheng, Zheng, Yu, Gonzalez, Zhang,
  and Stoica]{kwon2023efficient}
Woosuk Kwon, Zhuohan Li, Siyuan Zhuang, Ying Sheng, Lianmin Zheng, Cody~Hao Yu,
  Joseph~E. Gonzalez, Hao Zhang, and Ion Stoica.
\newblock Efficient memory management for large language model serving with
  pagedattention.
\newblock In \emph{Proceedings of the ACM SIGOPS 29th Symposium on Operating
  Systems Principles}, 2023.

\bibitem[Lewkowycz et~al.(2022)Lewkowycz, Andreassen, Dohan, Dyer, Michalewski,
  Ramasesh, Slone, Anil, Schlag, Gutman-Solo, et~al.]{lewkowycz2022solving}
Aitor Lewkowycz, Anders Andreassen, David Dohan, Ethan Dyer, Henryk
  Michalewski, Vinay Ramasesh, Ambrose Slone, Cem Anil, Imanol Schlag, Theo
  Gutman-Solo, et~al.
\newblock Solving quantitative reasoning problems with language models.
\newblock \emph{Advances in Neural Information Processing Systems},
  35:\penalty0 3843--3857, 2022.

\bibitem[Li et~al.(2024)Li, Liu, Zhou, and Ma]{Li2024ChainOT}
Zhiyuan Li, Hong Liu, Denny Zhou, and Tengyu Ma.
\newblock Chain of thought empowers transformers to solve inherently serial
  problems.
\newblock \emph{ArXiv}, abs/2402.12875, 2024.

\bibitem[Li et~al.(2025)Li, Zhang, Zhang, Zhang, Liu, Yao, Xu, Zheng, Wang,
  Chen, Zhang, Yin, Dong, Guo, Song, and Liu]{li202512surveyreasoning}
Zhong-Zhi Li, Duzhen Zhang, Ming-Liang Zhang, Jiaxin Zhang, Zengyan Liu, Yuxuan
  Yao, Haotian Xu, Junhao Zheng, Pei-Jie Wang, Xiuyi Chen, Yingying Zhang, Fei
  Yin, Jiahua Dong, Zhijiang Guo, Le~Song, and Cheng-Lin Liu.
\newblock From system 1 to system 2: A survey of reasoning large language
  models, 2025.
\newblock URL \url{https://arxiv.org/abs/2502.17419}.

\bibitem[Liang et~al.(2024)Liang, Song, Zheng, Wang, Yu, Li, Li, Xiong, and
  Li]{liang2024internalconsistencyselffeedbacklarge}
Xun Liang, Shichao Song, Zifan Zheng, Hanyu Wang, Qingchen Yu, Xunkai Li,
  Rong-Hua Li, Feiyu Xiong, and Zhiyu Li.
\newblock Internal consistency and self-feedback in large language models: A
  survey.
\newblock \emph{CoRR}, abs/2407.14507, 2024.
\newblock URL \url{https://doi.org/10.48550/arXiv.2407.14507}.

\bibitem[Lightman et~al.(2023)Lightman, Kosaraju, Burda, Edwards, Baker, Lee,
  Leike, Schulman, Sutskever, and Cobbe]{lightman2023let}
Hunter Lightman, Vineet Kosaraju, Yuri Burda, Harrison Edwards, Bowen Baker,
  Teddy Lee, Jan Leike, John Schulman, Ilya Sutskever, and Karl Cobbe.
\newblock Let's verify step by step.
\newblock In \emph{The Twelfth International Conference on Learning
  Representations}, 2023.

\bibitem[Lobo et~al.(2025)Lobo, Agarwal, and Lakkaraju]{lobo-etal-2025-impact}
Elita Lobo, Chirag Agarwal, and Himabindu Lakkaraju.
\newblock On the impact of fine-tuning on chain-of-thought reasoning.
\newblock In Luis Chiruzzo, Alan Ritter, and Lu~Wang (eds.), \emph{Proceedings
  of the 2025 Conference of the Nations of the Americas Chapter of the
  Association for Computational Linguistics: Human Language Technologies
  (Volume 1: Long Papers)}, pp.\  11679--11698, Albuquerque, New Mexico, April
  2025. Association for Computational Linguistics.
\newblock ISBN 979-8-89176-189-6.
\newblock URL \url{https://aclanthology.org/2025.naacl-long.584/}.

\bibitem[MacQueen(1967)]{macqueen1967some}
James MacQueen.
\newblock Some methods for classification and analysis of multivariate
  observations.
\newblock In \emph{Proceedings of the Fifth Berkeley Symposium on Mathematical
  Statistics and Probability, Volume 1: Statistics}, volume~5, pp.\  281--298.
  University of California press, 1967.

\bibitem[Moritz et~al.(2018)Moritz, Nishihara, Wang, Tumanov, Liaw, Liang,
  Elibol, Yang, Paul, Jordan, et~al.]{moritz2018ray}
Philipp Moritz, Robert Nishihara, Stephanie Wang, Alexey Tumanov, Richard Liaw,
  Eric Liang, Melih Elibol, Zongheng Yang, William Paul, Michael~I Jordan,
  et~al.
\newblock Ray: A distributed framework for emerging $\{$AI$\}$ applications.
\newblock In \emph{13th USENIX symposium on operating systems design and
  implementation (OSDI 18)}, pp.\  561--577, 2018.

\bibitem[{OpenAI}({\natexlab{a}})]{openai_o1}
{OpenAI}.
\newblock Introducing openai o1, {\natexlab{a}}.
\newblock URL \url{https://openai.com/o1/}.

\bibitem[{OpenAI}({\natexlab{b}})]{openai_o3_o4}
{OpenAI}.
\newblock Introducing openai o3 and o4-mini, {\natexlab{b}}.
\newblock URL \url{https://openai.com/index/introducing-o3-and-o4-mini/}.

\bibitem[Qwen()]{qwen3}
Qwen.
\newblock Qwen3: Think deeper, act faster.
\newblock URL \url{https://qwenlm.github.io/blog/qwen3/}.

\bibitem[Qwen(2025)]{qwq32b}
Qwen.
\newblock Qwq-32b: Embracing the power of reinforcement learning, March 2025.
\newblock URL \url{https://qwenlm.github.io/blog/qwq-32b/}.

\bibitem[Qwen et~al.(2025)Qwen, :, Yang, Yang, Zhang, Hui, Zheng, Yu, Li, Liu,
  Huang, Wei, Lin, Yang, Tu, Zhang, Yang, Yang, Zhou, Lin, Dang, Lu, Bao, Yang,
  Yu, Li, Xue, Zhang, Zhu, Men, Lin, Li, Tang, Xia, Ren, Ren, Fan, Su, Zhang,
  Wan, Liu, Cui, Zhang, and Qiu]{qwen2025qwen25technicalreport}
Qwen, :, An~Yang, Baosong Yang, Beichen Zhang, Binyuan Hui, Bo~Zheng, Bowen Yu,
  Chengyuan Li, Dayiheng Liu, Fei Huang, Haoran Wei, Huan Lin, Jian Yang,
  Jianhong Tu, Jianwei Zhang, Jianxin Yang, Jiaxi Yang, Jingren Zhou, Junyang
  Lin, Kai Dang, Keming Lu, Keqin Bao, Kexin Yang, Le~Yu, Mei Li, Mingfeng Xue,
  Pei Zhang, Qin Zhu, Rui Men, Runji Lin, Tianhao Li, Tianyi Tang, Tingyu Xia,
  Xingzhang Ren, Xuancheng Ren, Yang Fan, Yang Su, Yichang Zhang, Yu~Wan,
  Yuqiong Liu, Zeyu Cui, Zhenru Zhang, and Zihan Qiu.
\newblock Qwen2.5 technical report, 2025.
\newblock URL \url{https://arxiv.org/abs/2412.15115}.

\bibitem[Rein et~al.(2024)Rein, Hou, Stickland, Petty, Pang, Dirani, Michael,
  and Bowman]{rein2024gpqa}
David Rein, Betty~Li Hou, Asa~Cooper Stickland, Jackson Petty, Richard~Yuanzhe
  Pang, Julien Dirani, Julian Michael, and Samuel~R Bowman.
\newblock Gpqa: A graduate-level google-proof q\&a benchmark.
\newblock In \emph{First Conference on Language Modeling}, 2024.

\bibitem[Schulman et~al.(2017)Schulman, Wolski, Dhariwal, Radford, and
  Klimov]{schulman2017proximalpolicyoptimizationalgorithms}
John Schulman, Filip Wolski, Prafulla Dhariwal, Alec Radford, and Oleg Klimov.
\newblock Proximal policy optimization algorithms, 2017.
\newblock URL \url{https://arxiv.org/abs/1707.06347}.

\bibitem[Shao et~al.(2023)Shao, Gong, Shen, Huang, Duan, and
  Chen]{Shao2023SyntheticPG}
Zhihong Shao, Yeyun Gong, Yelong Shen, Minlie Huang, Nan Duan, and Weizhu Chen.
\newblock Synthetic prompting: generating chain-of-thought demonstrations for
  large language models.
\newblock In \emph{Proceedings of the 40th International Conference on Machine
  Learning}, ICML'23. JMLR.org, 2023.

\bibitem[Shao et~al.(2024)Shao, Wang, Zhu, Xu, Song, Bi, Zhang, Zhang, Li, Wu,
  and Guo]{shao2024deepseekmathpushinglimitsmathematical}
Zhihong Shao, Peiyi Wang, Qihao Zhu, Runxin Xu, Junxiao Song, Xiao Bi, Haowei
  Zhang, Mingchuan Zhang, Y.~K. Li, Y.~Wu, and Daya Guo.
\newblock Deepseekmath: Pushing the limits of mathematical reasoning in open
  language models, 2024.
\newblock URL \url{https://arxiv.org/abs/2402.03300}.

\bibitem[Silver et~al.(2017)Silver, Hubert, Schrittwieser, Antonoglou, Lai,
  Guez, Lanctot, Sifre, Kumaran, Graepel, Lillicrap, Simonyan, and
  Hassabis]{silver2017masteringchessshogiselfplay}
David Silver, Thomas Hubert, Julian Schrittwieser, Ioannis Antonoglou, Matthew
  Lai, Arthur Guez, Marc Lanctot, Laurent Sifre, Dharshan Kumaran, Thore
  Graepel, Timothy Lillicrap, Karen Simonyan, and Demis Hassabis.
\newblock Mastering chess and shogi by self-play with a general reinforcement
  learning algorithm, 2017.
\newblock URL \url{https://arxiv.org/abs/1712.01815}.

\bibitem[THUDM(2025)]{thudm_t1}
THUDM.
\newblock T1, 2025.
\newblock URL \url{https://huggingface.co/datasets/THUDM/T1}.

\bibitem[Trung et~al.(2024)Trung, Zhang, Jie, Sun, Jin, and
  Li]{trung-etal-2024-reft}
Luong Trung, Xinbo Zhang, Zhanming Jie, Peng Sun, Xiaoran Jin, and Hang Li.
\newblock {R}e{FT}: Reasoning with reinforced fine-tuning.
\newblock In Lun-Wei Ku, Andre Martins, and Vivek Srikumar (eds.),
  \emph{Proceedings of the 62nd Annual Meeting of the Association for
  Computational Linguistics (Volume 1: Long Papers)}, pp.\  7601--7614,
  Bangkok, Thailand, August 2024. Association for Computational Linguistics.
\newblock \doi{10.18653/v1/2024.acl-long.410}.
\newblock URL \url{https://aclanthology.org/2024.acl-long.410/}.

\bibitem[Wang et~al.(2024{\natexlab{a}})Wang, Qin, Jacobs, Wu, Holmes, Yao,
  Rajbhandari, Ruwase, Yan, Yang, et~al.]{wang2024zero++}
Guanhua Wang, Heyang Qin, Sam~Ade Jacobs, Xiaoxia Wu, Connor Holmes, Zhewei
  Yao, Samyam Rajbhandari, Olatunji Ruwase, Feng Yan, Lei Yang, et~al.
\newblock Zero++: Extremely efficient collective communication for large model
  training.
\newblock In \emph{The Twelfth International Conference on Learning
  Representations}, 2024{\natexlab{a}}.

\bibitem[Wang et~al.(2024{\natexlab{b}})Wang, Ma, Meng, Qin, Shen, Zhang, Wu,
  Liu, Bian, Xu, Wang, and Zhao]{wang2024steponfeet}
Haoyu Wang, Guozheng Ma, Ziqiao Meng, Zeyu Qin, Li~Shen, Zhong Zhang, Bingzhe
  Wu, Liu Liu, Yatao Bian, Tingyang Xu, Xueqian Wang, and Peilin Zhao.
\newblock Step-on-feet tuning: Scaling self-alignment of {LLM}s via
  bootstrapping.
\newblock In \emph{ICML 2024 Workshop on Models of Human Feedback for AI
  Alignment}, 2024{\natexlab{b}}.
\newblock URL \url{https://openreview.net/forum?id=lAXNiTcMar}.

\bibitem[Wang et~al.(2023{\natexlab{a}})Wang, Wei, Schuurmans, Le, Chi, Narang,
  Chowdhery, and Zhou]{wang2023selfconsistency}
Xuezhi Wang, Jason Wei, Dale Schuurmans, Quoc~V Le, Ed~H. Chi, Sharan Narang,
  Aakanksha Chowdhery, and Denny Zhou.
\newblock Self-consistency improves chain of thought reasoning in language
  models.
\newblock In \emph{The Eleventh International Conference on Learning
  Representations}, 2023{\natexlab{a}}.
\newblock URL \url{https://openreview.net/forum?id=1PL1NIMMrw}.

\bibitem[Wang et~al.(2023{\natexlab{b}})Wang, Kordi, Mishra, Liu, Smith,
  Khashabi, and Hajishirzi]{wang2023selfinstructaligninglanguagemodels}
Yizhong Wang, Yeganeh Kordi, Swaroop Mishra, Alisa Liu, Noah~A. Smith, Daniel
  Khashabi, and Hannaneh Hajishirzi.
\newblock Self-instruct: Aligning language models with self-generated
  instructions.
\newblock In Anna Rogers, Jordan Boyd-Graber, and Naoaki Okazaki (eds.),
  \emph{Proceedings of the 61st Annual Meeting of the Association for
  Computational Linguistics (Volume 1: Long Papers)}, pp.\  13484--13508,
  Toronto, Canada, July 2023{\natexlab{b}}. Association for Computational
  Linguistics.
\newblock \doi{10.18653/v1/2023.acl-long.754}.
\newblock URL \url{https://aclanthology.org/2023.acl-long.754/}.

\bibitem[Wei et~al.(2022)Wei, Wang, Schuurmans, Bosma, Xia, Chi, Le, Zhou,
  et~al.]{wei2022chain}
Jason Wei, Xuezhi Wang, Dale Schuurmans, Maarten Bosma, Fei Xia, Ed~Chi, Quoc~V
  Le, Denny Zhou, et~al.
\newblock Chain-of-thought prompting elicits reasoning in large language
  models.
\newblock \emph{Advances in neural information processing systems},
  35:\penalty0 24824--24837, 2022.

\bibitem[Wolf et~al.(2020)Wolf, Debut, Sanh, Chaumond, Delangue, Moi, Cistac,
  Rault, Louf, Funtowicz, et~al.]{wolf2020transformers}
Thomas Wolf, Lysandre Debut, Victor Sanh, Julien Chaumond, Clement Delangue,
  Anthony Moi, Pierric Cistac, Tim Rault, R{\'e}mi Louf, Morgan Funtowicz,
  et~al.
\newblock Transformers: State-of-the-art natural language processing.
\newblock In \emph{Proceedings of the 2020 conference on empirical methods in
  natural language processing: system demonstrations}, pp.\  38--45, 2020.

\bibitem[Wu et~al.(2024)Wu, Jiang, Xiong, Ruan, Ding, Guo, Wen, Zhou, and
  Deng]{wu2024hummer}
Yusen Wu, Li~Jiang, Junwu Xiong, Jingqing Ruan, Yichuan Ding, Qingpei Guo,
  Zujie Wen, Jun Zhou, and Xiaotie Deng.
\newblock Hummer: Towards limited competitive preference dataset.
\newblock In \emph{Proceedings of the Conference on Learning for Language
  Modeling (COLM)}, 2024.
\newblock URL \url{https://openreview.net/forum?id=aKwQPRjdGa}.
\newblock Under review.

\bibitem[Xu et~al.(2025{\natexlab{a}})Xu, Peng, Awadalla, Chen, Chen, Gao, Kim,
  Li, Ren, Shen, Wang, Xu, Gao, and
  Chen]{xu2025phi4minireasoningexploringlimitssmall}
Haoran Xu, Baolin Peng, Hany Awadalla, Dongdong Chen, Yen-Chun Chen, Mei Gao,
  Young~Jin Kim, Yunsheng Li, Liliang Ren, Yelong Shen, Shuohang Wang, Weijian
  Xu, Jianfeng Gao, and Weizhu Chen.
\newblock Phi-4-mini-reasoning: Exploring the limits of small reasoning
  language models in math, 2025{\natexlab{a}}.
\newblock URL \url{https://arxiv.org/abs/2504.21233}.

\bibitem[Xu et~al.(2025{\natexlab{b}})Xu, Jiang, Niu, Deng, Poovendran, Choi,
  and Lin]{xu2025magpie}
Zhangchen Xu, Fengqing Jiang, Luyao Niu, Yuntian Deng, Radha Poovendran, Yejin
  Choi, and Bill~Yuchen Lin.
\newblock Magpie: Alignment data synthesis from scratch by prompting aligned
  {LLM}s with nothing.
\newblock In \emph{The Thirteenth International Conference on Learning
  Representations}, 2025{\natexlab{b}}.
\newblock URL \url{https://openreview.net/forum?id=Pnk7vMbznK}.

\bibitem[Yang et~al.(2024)Yang, Zhang, Hui, Gao, Yu, Li, Liu, Tu, Zhou, Lin,
  Lu, Xue, Lin, Liu, Ren, and
  Zhang]{yang2024qwen25mathtechnicalreportmathematical}
An~Yang, Beichen Zhang, Binyuan Hui, Bofei Gao, Bowen Yu, Chengpeng Li,
  Dayiheng Liu, Jianhong Tu, Jingren Zhou, Junyang Lin, Keming Lu, Mingfeng
  Xue, Runji Lin, Tianyu Liu, Xingzhang Ren, and Zhenru Zhang.
\newblock Qwen2.5-math technical report: Toward mathematical expert model via
  self-improvement, 2024.
\newblock URL \url{https://arxiv.org/abs/2409.12122}.

\bibitem[Yao et~al.(2023)Yao, Yu, Zhao, Shafran, Griffiths, Cao, and
  Narasimhan]{yao2023tree}
Shunyu Yao, Dian Yu, Jeffrey Zhao, Izhak Shafran, Tom Griffiths, Yuan Cao, and
  Karthik Narasimhan.
\newblock Tree of thoughts: Deliberate problem solving with large language
  models.
\newblock \emph{Advances in neural information processing systems},
  36:\penalty0 11809--11822, 2023.

\bibitem[Yeo et~al.(2025)Yeo, Tong, Niu, Neubig, and
  Yue]{yeo2025demystifyinglongchainofthoughtreasoning}
Edward Yeo, Yuxuan Tong, Morry Niu, Graham Neubig, and Xiang Yue.
\newblock Demystifying long chain-of-thought reasoning in llms, 2025.
\newblock URL \url{https://arxiv.org/abs/2502.03373}.

\bibitem[Zheng et~al.(2024)Zheng, Yin, Xie, Sun, Huang, Yu, Cao, Kozyrakis,
  Stoica, Gonzalez, Barrett, and Sheng]{SGLang}
Lianmin Zheng, Liangsheng Yin, Zhiqiang Xie, Chuyue Sun, Jeff Huang, Cody~Hao
  Yu, Shiyi Cao, Christos Kozyrakis, Ion Stoica, Joseph~E. Gonzalez, Clark
  Barrett, and Ying Sheng.
\newblock Sglang: Efficient execution of structured language model programs.
\newblock In A.~Globerson, L.~Mackey, D.~Belgrave, A.~Fan, U.~Paquet,
  J.~Tomczak, and C.~Zhang (eds.), \emph{Advances in Neural Information
  Processing Systems}, volume~37, pp.\  62557--62583. Curran Associates, Inc.,
  2024.
\newblock URL
  \url{https://proceedings.neurips.cc/paper_files/paper/2024/file/724be4472168f31ba1c9ac630f15dec8-Paper-Conference.pdf}.

\end{thebibliography}
\bibliographystyle{sharp}

\clearpage
\appendix

\section{SHARP Self-Alignment Strategy Constraints}

\begin{tcolorbox}[colback=gray!5, colframe=gray!80!black, title=SHARP Self-Alignment Strategy Constraints Details]
\textbf{Problem Difficulty \& Thematic Diversity Alignment}: Generate highly complex problems (graduate- or Olympiad-level) covering a wide range of STEM topics, covering expert-level AI themes. Difficulty is benchmarked against top exams and datasets (GPQA, etc.). Thematic coverage uses role-playing prompts template and a three-tier subject-category-topic framework.

\medskip
\textbf{Logical Consistency Alignment}: Problem-solving must rely solely on rigorous reasoning or systematic derivation, avoiding pattern matching, heuristics, shortcuts, or fabrication. All intermediate steps require justification, preventing logical gaps or errors due to intuition.

\medskip
\textbf{Ground Truth \& Structure Alignment}: Answers must be single, verifiable numerical values (plain numbers, units, ratios, STEM formulas/equations). Avoid hard-to-verify formats (set operations, free text). For multi-solution problems, mandate a specific aggregation (e.g., sum or sum of squares, etc.) for a unique, objectively verifiable answer. Expand beyond single QA to include multi-solution problems (requiring summary values) (e.g., ``calculate total moles of all possible products'').

\medskip
\textbf{Problem Authenticity Alignment}: Problems should be novel, based on authoritative knowledge, but not directly copied. They must be unambiguous, unbiased, accurate, and internally consistent, avoiding nonsensical or hallucinated scenarios.

\medskip
\textbf{Language Consistency Alignment}: The entire generation process (problem statement, reasoning method, solution presentation) must use a single language (e.g., English or Chinese) to prevent multilingual confusion leading to reasoning errors or bad verification cases.

\medskip
\textbf{Problem Structure Consistency Alignment}: Problems must contain only a single primary question, avoiding sub-questions, derivatives, or branching logic that leads to unverifiable cases.

\medskip
\textbf{Modality Consistency Alignment}: Problems must be strictly text-based, describing any necessary complex structures (e.g., chemical molecules, genetic diagrams) textually.

\medskip
\textbf{Formatting Alignment}: Use specific delimiters (e.g., \texttt{<question\_start>}, \texttt{<question\_end>}) for the problem statement and a standardized format (e.g., \texttt{\textbackslash boxed\{\{\$answer\}\}}) for the final answer.
\end{tcolorbox}

The complete template including the \textbf{SHARP} self-alignment strategy constraints for constructing the challenge problem is shown in the following Table \ref{tab:sharp_prompt} and can also be found via this \href{https://anonymous.4open.science/r/sharp-93F6/sharp_prompt_template.jsonl}{link}.

\section{Performance Analysis of Distilling and RL Zero Model Reasoning Training with SHARP Samples}
\subsection{Distilling Training Model Performance Analysis}
\begin{figure}
    \centering
    \includegraphics[width=1\linewidth]{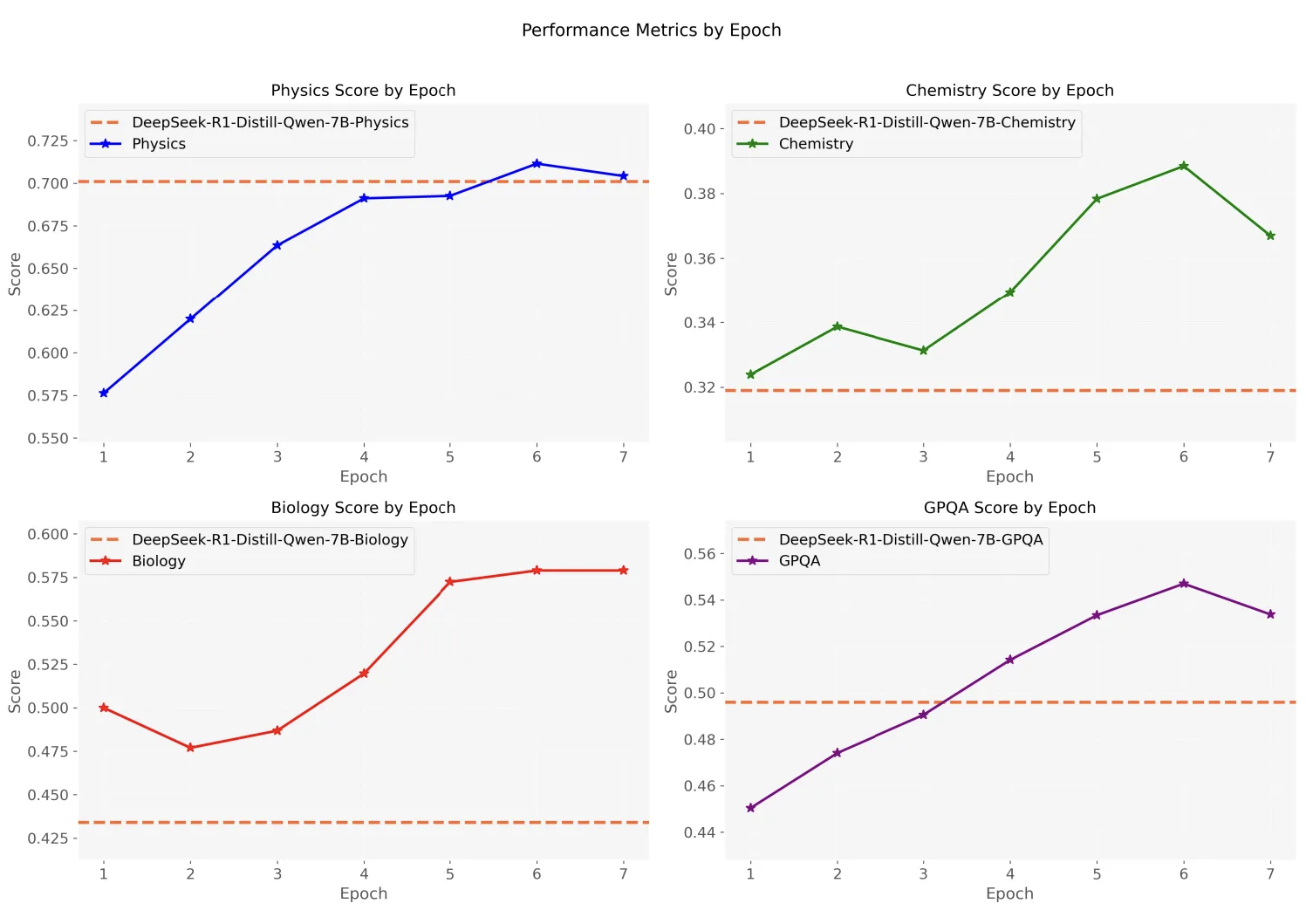}
    \caption{GPQA score improvement of single STEM disciplines (physics, chemistry, and biology) of \textbf{SHARP-Qwen2.5-7B-Instruct-Distill} relative to benchmark model \textbf{DeepSeek-R1-Distill-Qwen-7B} in overall ablation of STEM data generated by the \textbf{SHARP} approach and fused with some open-source mathematical data. (The $x$-axis represents the different epochs run during the training of the distill models, and the $y$-axis represents the GPQA score evaluation results corresponding to the checkpoints of the models generated at different epochs).}
    \label{fig:sharp_distill_gpqa_compare}
\end{figure}

Fig.\ref{fig:sharp_distill_gpqa_compare} compares models trained on \textbf{SHARP-augmented Qwen2.5-7B-Instruct-Distill (Baseline)} and the strong benchmark \textbf{DeepSeek-R1-Distill-Qwen-7B} with samples generated by the \textbf{SHARP} approach  across three STEM subjects: physics, chemistry, and biology. In addition, the physics, chemistry, and biology subjects all had positive improvements, and the chemistry and biology subjects compared with the DeepSeek chemistry subject improved significantly, indicating the effectiveness of our designed \textbf{SHARP} self-alignment strategy and reasoning training model, reflecting the improvement of the model in general knowledge and reasoning ability.

\begin{figure}
    \centering
    \includegraphics[width=1\linewidth]{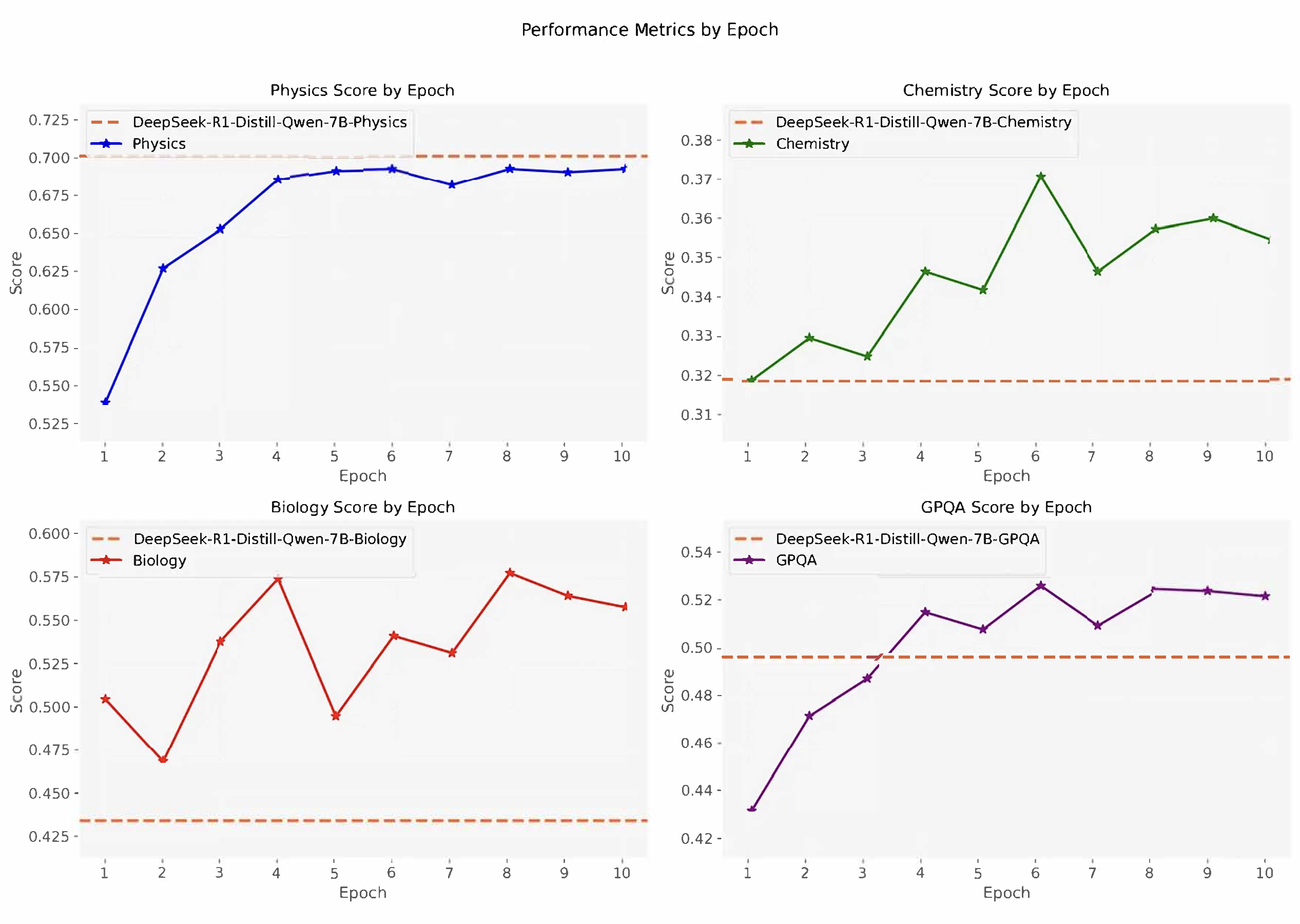}
    \caption{GPQA score improvement of single STEM disciplines (physics, chemistry, and biology) of \textbf{SHARP-Qwen2.5-7B-Instruct-Distill} relative to benchmark model \textbf{DeepSeek-R1-Distill-Qwen-7B} in ablation of STEM data generated by the \textbf{SHARP} approach. (The meanings of the $x$ and $y$ axes are the same as those in Fig.\ref{fig:sharp_distill_gpqa_compare}.)}
    \label{fig:sharp_instruct_distill_gpqa_compare}
\end{figure}

\begin{figure}
    \centering
    \includegraphics[width=1\linewidth]{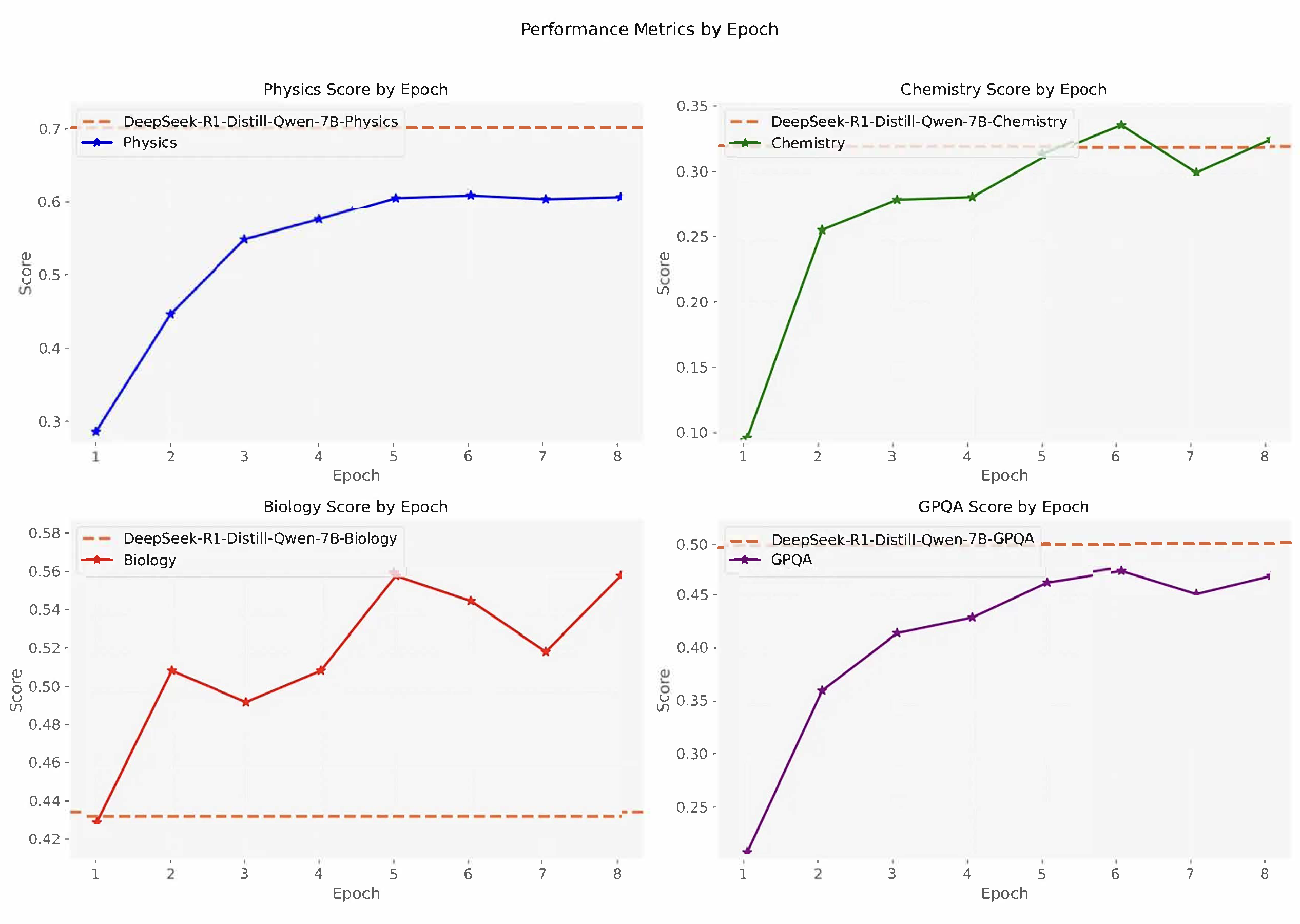}
    \caption{GPQA score improvement of single STEM disciplines (physics, chemistry, and biology) of \textbf{SHARP-Qwen2.5-7B-Instruct-Distill} relative to benchmark model \textbf{DeepSeek-R1-Distill-Qwen-7B} in ablation of some open-source mathematical data. (The meanings of the $x$ and $y$ axes are the same as those in Fig.\ref{fig:sharp_distill_gpqa_compare}.)}
    \label{fig:sharp_instruct_distill_math_compare}
\end{figure}

Also, model trained with \textbf{SHARP} problems only are significantly better than mathematical only distillation problems in improving the ability of physics, chemistry, and biology, as shown in \ref{fig:sharp_instruct_distill_gpqa_compare} and \ref{fig:sharp_instruct_distill_math_compare}, and thus significantly better than mathematical only distillation problems in the overall GPQA benchmark (Here, the mathematics data here accounts for 27.3\%, mainly from \citep{deepscaler, artofproblemsolving, thudm_t1}, mathematics competition problems from all over the world, well-known universities, etc.). As seen from the Fig.\ref{fig:sharp_instruct_distill_math_compare}, the GPQA score of the distillation model is not as significant in improving the chemistry index in the pure mathematics data set as in physics and biology. This also shows, to some extent, that the attributes of chemistry and mathematical reasoning are relatively different.

We conduct these distillation model supervised finetuning across 10 epochs for all datasets and a learning rate of 5e-6. We employ a cosine learning rate scheduler, ensuring that the final learning rate reaches 1\% of the peak value. We train these models with about 190,000 samples on 32 NVIDIA H800 GPUs for 10 hours. These core parameters for training are set as in Table \ref{table:distill_model_sft_parameters}:
\begin{table}[h!]
\centering
\begin{tabular}{lcc}
\toprule
\textbf{Parameter Name} & \textbf{Value} \\
\midrule
max\_length & 16384 \\
learning\_rate & 5e-6 \\
lr\_scheduler\_type & cosine \\
warmup\_ratio & 0.01 \\
\midrule
\end{tabular}
\caption{Distill Model Core Parameters.}
\label{table:distill_model_sft_parameters}
\end{table}

Specific challenging problem samples generated by the \textbf{SHARP} approach used to train distill models can be found via this \href{https://anonymous.4open.science/r/sharp-93F6/sharp_problems_distill.jsonl}{link}.

\subsection{RL Zero Training Model Performance Analysis}
\begin{figure}
    \centering
    \includegraphics[width=1\linewidth]{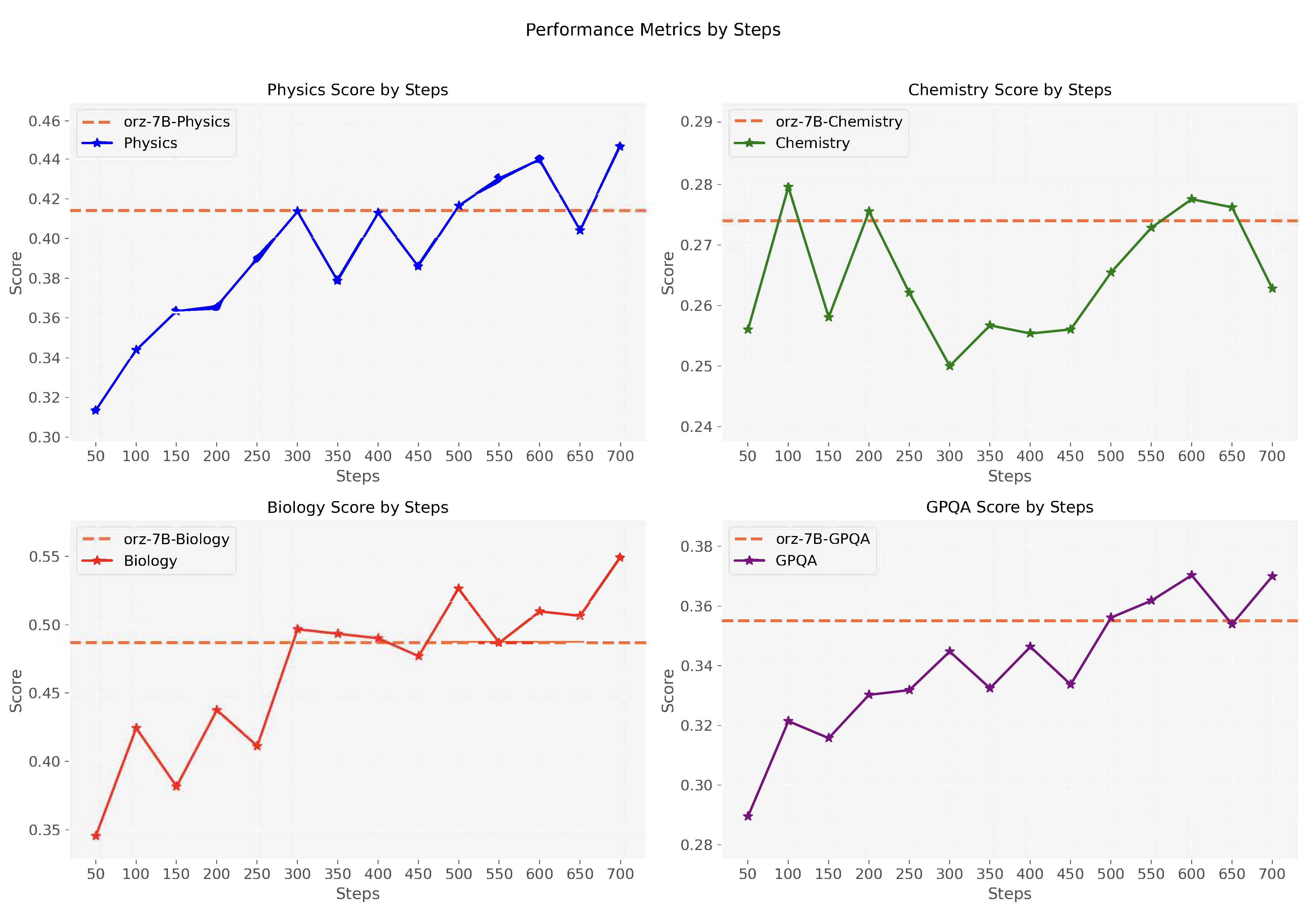}
    \caption{\textbf{Open-Reasoner-Zero-7B} performance in ablation of STEM data generated by the \textbf{SHARP} approach. (The $x$-axis represents the different running steps during the training of the reinforcement learning reasoning model, and the $y$-axis represents the GPQA score evaluation results corresponding to the checkpoints of the models generated at different steps.)}
    \label{fig:sharp_rl_zero_gpqa_compare}
\end{figure}

\begin{figure}
    \centering
    \includegraphics[width=1\linewidth]{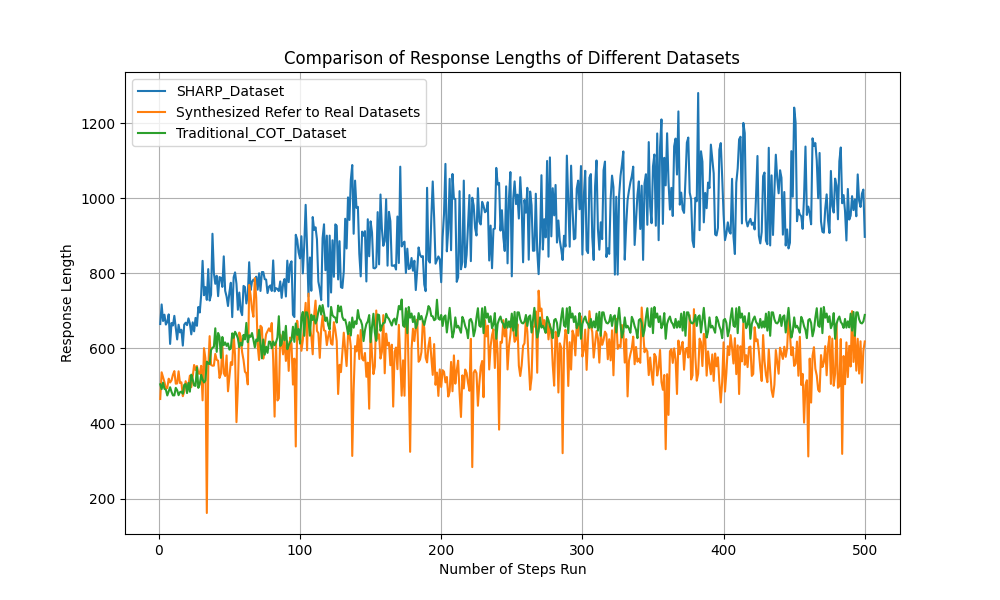}
    \caption{The response length of RL Zero model in ablation of chemistry data from three different problems datasets, 1) problems synthesized through traditional COT, 2) problems augmented synthesized referencing to real challenging chemistry exercises and 3) problems synthesized \textbf{SHARP} approach. (The $x$-axis represents the different running steps during the training of the reinforcement learning reasoning model, and the $y$-axis represents the response length when the models runs at corresponding steps.)}
    \label{fig:rl_response_length_chemistry}
\end{figure}

\begin{figure}
    \centering
    \includegraphics[width=1\linewidth]{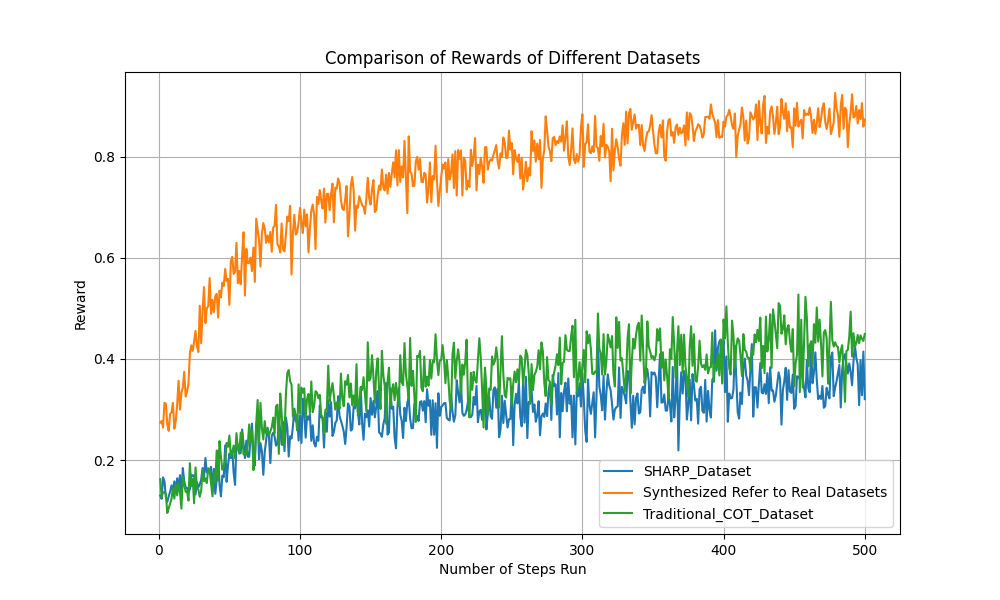}
    \caption{The reward RL Zero model in ablation of chemistry data from three different problems datasets, 1) problems synthesized through traditional COT, 2) problems augmented synthesized referencing to real challenging chemistry exercises and 3) problems synthesized \textbf{SHARP} approach. (The $x$-axis represents the different running steps during the training of the reinforcement learning reasoning model, and the $y$-axis represents the reward when the models runs at corresponding steps.)}
    \label{fig:rl_reward_chemistry}
\end{figure}

As shown in Table \ref{table:evaluation_7b} and Fig.\ref{fig:sharp_rl_zero_gpqa_compare}, after we added \textbf{SHARP} problems as the main training data (about 73\%) for RL Zero enhanced reasoning training in \textbf{SHARP-Open-Reasoner-Zero-7B}, it has exceeded the pure mathematics RL Zero mathematical reasoning model \textbf{Open-Reasoner-Zero-7B (Baseline)} by about 4.22\%, and the single subjects of physics and biology have exceeded the \textbf{Open-Reasoner-Zero-7B (Baseline)} model, and the chemistry subject is basically the same, which shows that the \textbf{SHARP} self-alignment strategy and inference training system implemented have improved the pure complex reasoning ability of the model. Especially for chemistry, we compare two key metrics for evaluation RL Zero training model: the response length (which usually is used to indicate the complexity of the problems), as shown in Fig.\ref{fig:rl_response_length_chemistry} and reward value (whose values are usually used to indicate the difficulty degree of the problems) as shown in Fig.\ref{fig:rl_reward_chemistry} in three different problems datasets, 1) problems synthesized through traditional COT, 2) problems augmented synthesized referencing to real challenging chemistry exercises and 3) problems synthesized \textbf{SHARP} approach. Through the experimental comparison of each stage, the difficulty of the sample problems generated by our \textbf{SHARP} approach has significantly increased the response length for the correct answer, and the distribution of rewards has shown a significant downward trend. Although the GPQA score of the chemistry subject has not improved significantly, through combined with the gradual and significant improvement of the experimental evaluation indicators, it demonstrates the effectiveness of the \textbf{SHARP} approach in improving the complex reasoning ability of the model, and also indicates that increasing the model's own complex problems to obtain the groundtruth can further significantly increase the effect of the model.

Specific challenging problems generated by the \textbf{SHARP} approach used to train RL Zero models are shown in the Table \ref{table:sharp_problem_phy_chem_bio}, and other samples can be found via this \href{https://anonymous.4open.science/r/sharp-93F6/sharp_problems_rl_zero.jsonl}{link}. 

\begin{table}[h!]
  \centering
  \begin{tabular}{p{5cm} p{9cm}} 
    \toprule
    \textbf{Subject Name} & \textbf{Problem and Reference Answer} \\
    \midrule
    Particle Physics‌, High Energy Physics & "problem": "Solve the following chemical problem step by step. The last line of your response should be of the form \verb+\boxed{$Answer$}+ (without quotes) where $Answer$ is the answer to the problem. A core-collapse supernova at a distance of 1 kiloparsec ($3.086 \times 10^{19}$ meters) releases $3 \times 10^{46}$ J of energy, with 99\% of this energy emitted as neutrinos. Each neutrino has an average energy of 10 MeV ($1.6 \times 10^{-12}$ J). A spherical lead detector with a radius of 10 meters is used to observe these neutrinos. Lead has a density of 11,340 kg/m$^3$ and an atomic mass of 207.2 g/mol. The neutrino-nucleus interaction cross section is $1 \times 10^{-43}$ cm$^2$ per nucleus. Assuming neutrinos are emitted isotropically and all physical quantities are uniform, calculate the total number of neutrino interactions in the detector. **Constants and Formulas:** \- Avogadro’s number: $N_A = 6.022 \times 10^{23} \, \text{mol}^{-1}$ \- Sphere volume: $V = \frac{4}{3} \pi r^3$ \- Neutrino flux at Earth: $\Phi = \frac{N_\nu}{4 \pi d^2}$ \- Interaction rate: $N_{\text{interactions}} = \Phi \cdot \sigma \cdot N_{\text{nuclei}}$ Please reason step by step, and put your final answer within \verb+\boxed{$Answer$}+.", "ref\_answer": "\boxed{2140}". \\
    Organic Chemistry & "problem": "Solve the following chemical problem step by step. The last line of your response should be of the form \verb+\boxed{$Answer$}+ (without quotes) where $Answer$ is the answer to the problem. An impure sample of zinc carbonate ($ZnCO_3$) undergoes thermal decomposition, releasing carbon dioxide gas. The mass loss due to $CO_2$ emission is measured as 2.64 g. The resulting zinc oxide (ZnO) is then reduced using excess carbon, producing 5.89 g of zinc metal. 1. Write the balanced equation for the decomposition of $ZnCO_3$. 2. Write the balanced equation for the carbon reduction of ZnO. 3. Determine the percentage purity of zinc in the original impure sample. Assume all reactions proceed to completion, and impurities do not participate in any reactions. (Atomic masses: Zn = 65.38 g/mol, C = 12.01 g/mol, O = 16.00 g/mol) Remember to put your final answer within \verb+\boxed{$Answer$}+", "ref\_answer": "\boxed{58.9\%}".\\
    Molecular Biology, Virology & "problem": "Solve the following biological problem step by step. The last line of your response should be of the form \verb+\boxed{$Answer$}+ (without quotes) where $Answer$ is the answer to the problem. The SARS-CoV-2 genome is a single-stranded RNA virus with a genome length of 29,903 nucleotides. The spike (S) protein gene constitutes 12.73\% of the genome. Each S protein monomer consists of amino acids with an average molecular weight of 110 Da. A single virion contains 2.5 femtograms (fg) of S protein. Calculate the total number of S protein trimers on the virion's surface. Use Avogadro's number ($6.022 \times 10^{23} \text{ mol}^{-1}$) for your calculations. Remember to put your final answer within \verb+\boxed{$Answer$}+", "ref\_answer": "\boxed{3586}".\\
    \bottomrule
  \end{tabular}
  \caption{The challenging problems of physics, chemistry, and biology generated by the \textbf{SHARP} approach.}
  \label{table:sharp_problem_phy_chem_bio}
\end{table}

To initialize RL Zero training, we employ the verified \textbf{SHARP}-generated dataset containing input questions and corresponding verified ground truth answers. We adopt Group Relative Policy Optimization (GRPO) \citep{shao2024deepseekmathpushinglimitsmathematical}, a memory-efficient reinforcement learning method well-suited to \textbf{SHARP}’s batch-verifiable training data. GRPO bypasses the need for a separate critic by estimating baselines from group-level sample scores. For each problem $q$, GRPO samples a group of outputs $o_1, o_2, \ldots, o_G$ from the old policy $\pi_{\theta_{\text{old}}}$ and then optimizes the policy model $\pi_\theta$ by maximizing the following objective:

\begin{align*}
\mathcal{J}_{\text{GRPO}}(\theta) = \mathbb{E} & \left[ q \sim P(Q), \{o_i\}_{i=1}^G \sim \pi_{\theta_{\text{old}}}(O|q) \right. \\
& \quad \times \frac{1}{G} \sum_{i=1}^G \frac{1}{|o_i|} \sum_{t=1}^{|o_i|} \left( \min \left( \frac{\pi_{\theta}(o_{i,t}|q, o_{i,<t})}{\pi_{\theta_{\text{old}}}(o_{i,t}|q, o_{i,<t})} \hat{A}_{i,t}, \text{clip}\left(\frac{\pi_{\theta}(o_{i,t}|q, o_{i,<t})}{\pi_{\theta_{\text{old}}}(o_{i,t}|q, o_{i,<t})}, 1-\epsilon, 1+\epsilon\right) \hat{A}_{i,t} \right) \right. \\
& \quad \left. \left. - \beta D_{\text{KL}}(\pi_{\theta}||\pi_{\text{ref}}) \right) \right] \\
D_{\text{KL}}[\pi_{\theta}||\pi_{\text{ref}}] ={}& \frac{\pi_{\text{ref}}(o_{i,t}|q, o_{i,<t})}{\pi_{\theta}(o_{i,t}|q, o_{i,<t})} - \log \frac{\pi_{\text{ref}}(o_{i,t}|q, o_{i,<t})}{\pi_{\theta}(o_{i,t}|q, o_{i,<t})} - 1. \quad (1)
\end{align*}

where $\epsilon$ and $\beta$ are hyperparameters, and $\hat{A}_{i,t}$ is the advantage, computed using a group of rewards $\{r_1, r_2, ..., r_G\}$ corresponding to the outputs within each group:
$$ \hat{A}_{i,t} = \frac{r_i - \text{mean}(\{r_1, r_2, ..., r_G\})}{\text{std}(\{r_1, r_2, ..., r_G\})} $$

Inspired by prior work \citep{deepseekai2025deepseekr1incentivizingreasoningcapability}, we employ a simplified reward function $\mathcal{R}\_{\text{acc}}$ grounded in binary accuracy. Unlike prior methods, \textbf{SHARP} enables reward assignment based solely on correctness, owing to its verified outputs. This design simplifies the reward signal while preserving alignment quality. Here, the accuracy reward $\mathcal{R}\_{\text{acc}}$ evaluates correctness based on whether the model's response $a_i$ is similar to the ground truth solution $a_i'$ to satisfy the correctness criteria:\\
\[
\mathcal{R}_{\text{acc}}(a_i, a_i') = \begin{cases} 1, & \text{if equal}(a_i, a_i'), \\ 0, & \text{otherwise.} \end{cases}
\]

We implement GRPO training on RL Zero models using OpenRLHF \citep{hu2024openrlhfeasytousescalablehighperformance}, an open-source RL framework built atop Ray \citep{moritz2018ray}, vLLM \citep{kwon2023efficient}, ZeRO-3 \citep{wang2024zero++}, and HuggingFace Transformers \citep{wolf2020transformers}. We train these models with about 190,000 samples on 256 NVIDIA H800 GPUs for 48 hours. Key algorithm parameters for RL Zero models training are set as in Table \ref{table:rl_algorithm_parameters}. For each prompt, we generate 64 diverse completions to support robust group-based reward estimation. The KL divergence constraint coefficient is fixed at 0.001 across all experiments. Additionally, we mix problems from various STEM domains during model training to ensure diverse learning. We report accuracy by averaging the results over greedy decoding across 16 independent inference runs, which ensures statistical stability while preserving inference consistency for the GPQA benchmark evaluation. Answers are extracted from the standardized \verb+\boxed{$Answer$}+ format to verify against the ground truth solutions to ensure correctness and ensure alignment with \textbf{SHARP}’s verifiability constraints during automatic evaluation.

\begin{table}[h!]
  \centering
  \begin{tabular}{@{}llp{10cm}@{}} 
    \toprule
    \textbf{Parameter Name} & \textbf{Value} & \textbf{Description} \\
    \midrule
    algorithm          & GRPO    & Reinforcement Learning Algorithm Used \\
    actor\_lr          & 1e-6    & Learning Rate of The Actor Network \\
    rollout\_bs        & 256     & Total Batch Size Used for Experience Collection \\
    train\_bs          & 16384   & Total Batch Size Used During Parameter Updates \\
    micro\_train\_bs   & 8       & Batch Size for a Single Forward Pass During Training \\
    micro\_rollout\_bs & 8       & Batch Size for a Single Forward Pass During Experience Collection \\
    sample\_k          & 64      & Number of output samples (G) generated per prompt by the policy for group reward estimation \\
    lambda             & 1.0     & Regularization Coefficient \\
    gamma              & 1.0     & Discount Factor \\
    kl                 & 0.001   & KL Divergence Constraint Coefficient \\
    max\_len           & 8192    & Maximum Sequence Length \\
    temperature        & 1.0     & Sampling Temperature \\
    \bottomrule
  \end{tabular}
  \caption{RL Zero Algorithm Core Parameters. Parameter values were tuned based on ablations to balance training stability, efficiency, and model performance.}
  \label{table:rl_algorithm_parameters}
\end{table}

These results affirm that \textbf{SHARP}-aligned training samples, combined with GRPO and rule-based accuracy rewards, significantly enhance RL Zero model performance in complex STEM reasoning tasks.

\section{SHARP Challenging Problem Datasets Analysis}
In this section, we first provide a detailed supplementary explanation of the overall dataflow process of the \textbf{SHARP} approach. Then we further analyze the coverage of the subject categories related to the data flow and the difficulty of the STEM challenging problems generated by the \textbf{SHARP} method based on this category.

\begin{figure}
    \centering
    \includegraphics[width=1\linewidth]{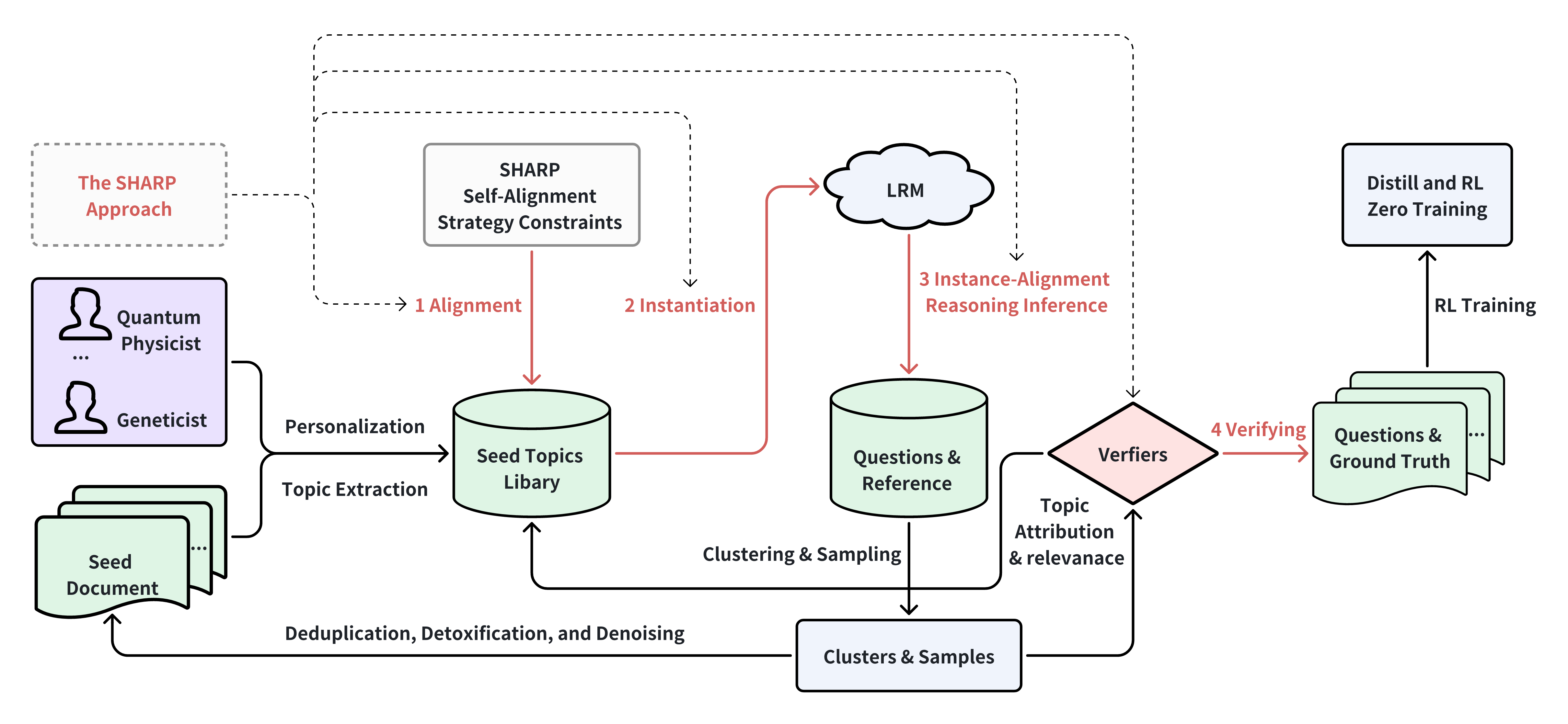}
    \caption{The overall dataflow process of the \textbf{SHARP} approach.}
    \label{fig:sharp_dataflow}
\end{figure}

The overall dataflow diagram of the construction of the \textbf{Seed Topics Library} and ``Three-Tier Category'' knowledge structure for STEM problems in the \textbf{SHARP} approach is shown in Fig.\ref{fig:sharp_dataflow}. As mentioned, they were mainly built by methods combined with the persona method \citep{ge2025scalingsyntheticdatacreation} and the Magpie-like method\citep{xu2025magpie} to generate a large number of personalized target topic query problems. Furthermore, in order to ensure the diversity and balance of the generated problems, a clustering strategy is designed, and these questions are distributed and balanced. In this way, we ensure that the generated problems cover a wide enough range of topics and have enough diversity under each topic, so as to provide comprehensive and balanced training problems for training LRMs.

\begin{figure}
    \centering
    \includegraphics[width=1\linewidth]{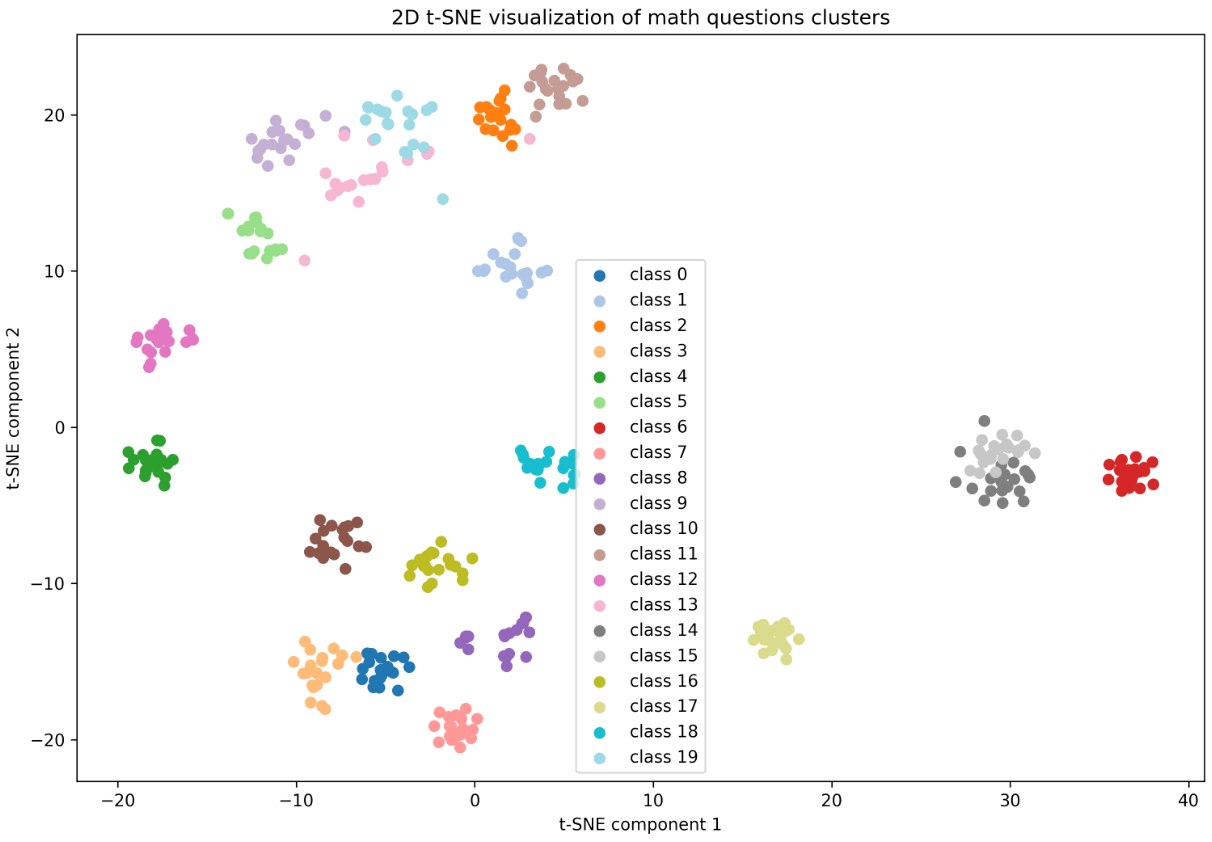}
    \caption{The K-means clustering results based on question embedding features extracted using BGE-m3 from problems generated by the \textbf{SHARP} approach.}
    \label{fig:clustering}
\end{figure}

Persona-driven \citep{ge2025scalingsyntheticdatacreation} prompts simulate domain experts with distinct problem-creation styles (e.g., a theoretical physicist vs. an organic chemist), ensuring varied problem framing and difficulty levels. Based on the persona method, we further improved the Magpie method to generate a large number of target topic query questions. We first designed a new ``Three-Tier Category'' knowledge structure with reference to the BISG category organization \citep{bisg_bisac} and subject characteristics to ensure that the first-level sub-disciplines, second-level self-disciplines, and basic concepts of each discipline are covered. Then we built high-quality seed documents to supplement and improve the themes of the ``Three-Tier Category'' knowledge structure of \textbf{SHARP} through the following two aspects. On the one hand, we analyzed and extracted topic keypoints based on high-quality open source training sets and question bankmarks such as \textbf{GPQA} (here, we only extracted topic keypoints without quoting or rewriting problems to prevent data leakage). On the other hand, we recall seed documents based on high-quality STEM textbooks, academic papers, Common Crawl, etc., and extract topics through the latest reasoning models such as Deepseek R1 and Qwen3 to obtain better topic diversity, thereby improving the model's generalization ability. Through the above series of methods, we ensure that the query problems have sufficient coverage while having expert persona characteristics, and at the same time ensure that the generated problems are consistent with the distribution of the current benchmark but have sufficient diversity and depth, so as to provide comprehensive, rich and challenging problems training dataset support for the complex reasoning of LLMs. In addition, we use BGE-m3 \citep{chen2024bgem3embeddingmultilingualmultifunctionality} to extract embedding features from the generated problems, and then use the K-means \citep{macqueen1967some} algorithm for clustering. We specify about 1,000 clusters via elbow method analysis on BGE-m3 embeddings to ensure that the number of clusters can cover most of the query problems, while ensuring that the queries within each cluster have a certain similarity and that there is sufficient difference between clusters. While clustering, each class is uniformly sampled to ensure the class balance of samples, and then an appropriate number of samples is extracted from each class for training. Fig. \ref{fig:clustering} illustrates the clustering results based on query embeddings, where we visualize a representative subset of 20 clusters. Each cluster exhibits strong intra-cluster cohesion, with samples tightly grouped in the embedding space. This suggests that queries within the same cluster share high semantic similarity. Moreover, clusters are well-separated from one another, indicating low semantic overlap across different groups. The clear inter-cluster boundaries highlight the effectiveness of our clustering pipeline in capturing meaningful semantic distinctions. Finally, the clustering and sampling results are processed for data deduplication, detoxification, and decontamination. Specific examples of the ``Three-Tier Category'' knowledge structure in the \textbf{Seed Topics Library} are shown in Table \ref{table:sharp_3_tier_cases}. 

\begin{table}[h!]
  \centering
  \begin{tabular}{p{4cm} p{4cm} p{5cm}} 
    \toprule
    \textbf{First-level Discipline} & \textbf{Second-level Discipline} & \textbf{Basic Knowledge-points} \\
    \midrule
    Theoretical Physics, High Energy Physics & Quantum Mechanics, Particle Physics & Energy levels, Heisenberg uncertainty principle, Lifetime-energy uncertainty relation, Energy resolution, Quantum states \\
    Organic Chemistry & Stereochemistry & Hydrogenation, Epoxidation, Nucleophilic substitution, Esterification, Limonene, Peracids, DCC coupling \\
    Basic Biology & Molecular Biology, Cancer Biology, Genetics, Epigenetics & Tumor suppressor genes, Gene expression, Epigenetic regulation, Gene silencing, Mouse models, Cancer cells \\
    \bottomrule
  \end{tabular}
  \caption{The ``Three-Tier Category'' structure examples of physics, chemistry and biology in the \textbf{SHARP} \textbf{Seed Topics Library}.}
  \label{table:sharp_3_tier_cases}
\end{table}

The ``Three-Tier Category'' structure, integrated with persona-driven prompts and clustering, ensures thematic diversity and logical consistency in SHARP-generated problems, directly contributing to enhanced model performance. Based on these data flow processing, the \textbf{SHARP} approach first combines the self-alignment strategy as shown in Algorithm \ref{alg:SHARP} to generate problems that help guide the reinforcement reasoning model at multiple levels. By integrating the persona \citep{ge2025scalingsyntheticdatacreation} role, the ``Three-Tier Category'' structure, and the \textbf{SHARP}  self-alignment strategy, the following template for creating problems is designed, as shown in the Table \ref{tab:sharp_prompt}. Then, the problem template is fused with the actual ``Three-Tier Category'' structure and knowledge framework through instantiation reasoning, thereby generating complex reasoning problems with self-alignment conditions and their corresponding reference answers. These complex questions and reference answers are then further verified, and finally, a high-quality, challenging question set and ground truth for complex reasoning are generated. The generated problems are not only conducive in characteristic disciplines and enhancing the generalization reasoning capabilities, but also generating difficult and logically consistent problems and corresponding verifiable answers that are conducive to LRMs through reinforcement learning via verifiable rewards (RLVR).

To enable scalable generation of \textbf{SHARP}-aligned STEM reasoning problems, we deploy the DeepSeek R1 model using the SGLang inference framework \citep{SGLang}, selected for its high-throughput serving capabilities, long-context support, and compatibility with structured output formatting.

This deployment is integrated into the \textbf{SHARP} \textbf{Instantiation} and \textbf{Inference} phases, where DeepSeek R1 is queried using templated prompts instantiated from persona roles, seed topics, and self-alignment constraints. Table~\ref{table:sglang_parameters} summarizes the server and inference configurations. Each request follows the \textbf{SHARP} prompting strategy (Table~\ref{tab:sharp_prompt}), with the \texttt{messages} field encoding the relevant persona roles, subject categories, and alignment directives.

The system, deployed on 8 NVIDIA H20 GPUs, supports 16K-token contexts and accommodates up to 48 concurrent requests, enabling efficient generation of high-difficulty, verifiable reasoning problems across diverse STEM domains. In total, 229{,}452 question–answer pairs were generated over 168 hours and subsequently evaluated through the \textbf{SHARP} \textbf{Verifying} phase for quality assurance prior to integration into RL Zero training.

\begin{table}[h!]
  \centering
  \begin{tabular}{@{}llp{10cm}@{}} 
    \toprule
    \textbf{Parameter Name} & \textbf{Value} & \textbf{Description} \\
    \midrule
    tp          & 8    & Tensor Parallelism \\
    max-running-requests & 48    & Concurrency \\
    mem-fraction-static & 0.94    & Memory Allocation \\
    context-length & 16384    & Context Length \\
    temperature  & 0.6   & Temperature \\
    max\_tokens & 4192 & Maximum Output Tokens  \\
    repetition\_penalty  & 1.05   & Repetition Penalty \\
    top\_p  & 0.8   & Top-p Sampling \\
    \bottomrule
  \end{tabular}
  \caption{Key SGLang server and inference parameters for each \textbf{SHARP} instantiated problem generation prompt.}
  \label{table:sglang_parameters}
\end{table}

\begin{table}
\begin{tcolorbox}[colback=gray!5, colframe=gray!80!black,title=The SHARP Approach Prompt]
\textless{}\textbf{Role\_Start}\textgreater{} \\
To test the \textless{} \textbf{Subject\_Name:} \textcolor{blue}{\{\textbf{subject}\_\textbf{name}\}}\textgreater{} reasoning and complex problem-solving skills of talented graduate students across various \textless{}\textbf{Subject\_Name:}\textcolor{blue}{\{\textbf{subject}\_\textbf{name}\}}\textgreater{} disciplines, you, a \textless{}\textbf{Persona\_Role}: \textcolor{red}{\{\textbf{persona}\_\textbf{role}\}}   \textgreater{} at a world-renowned institution, are creating a graduate- or Olympic-level challenging problem. \\
\textless{}\textbf{Role}\_\textbf{End}\textgreater{}

\textless{}\textbf{Task}\_\textbf{Description}\_\textbf{Start}\textgreater{} 
    \begin{itemize}[leftmargin=1.5em]
        \item You MUST refer to the following resources: \textless{}\textbf{SUB}\textgreater{} \textbf{Subject\_Name:} \textcolor{blue}{\{\textbf{subject}\_\textbf{name}\}} \textbf{Subdisciplines}: \textcolor{blue}{\{\textbf{subdisciplines}\}}\textless{}\textbf{SUB}\textgreater{}, \textless{}\textbf{BC}\textgreater{}\textbf{Basic Concepts}: \textcolor{blue}{{\textbf{\{basic\_concepts\}}}}\textless{}\textbf{BC}\textgreater{}. 
        \item You MUST randomly choose one or more items from the \textless{}\textbf{SUB}\textgreater{} \textbf{Subject\_Name:} \textcolor{blue}{\{\textbf{subject}\_\textbf{name}\}} \textbf{Subdisciplines}: \textcolor{blue}{\{\textbf{subdisciplines}\}}\textless{}\textbf{SUB}\textgreater{}, and then select several related concepts from the \textless{}\textbf{BC}\textgreater{}\textbf{Basic Concepts: \textcolor{blue}{{\textbf{\{basic\_concepts\}}}}}\textless{}\textbf{BC}\textgreater{} according to the subdisciplines to form an outline for the problem. Finally, create a calculation problem. 
    \end{itemize}
\textless{}\textbf{Task}\_\textbf{Description}\_\textbf{End}\textgreater{}

\textless{}\textbf{Requirements}\_\textbf{and}\_\textbf{Expectations}\_\textbf{Start}\textgreater{}\\
\textcolor{teal}{\text{Note: The problem must satisfy the following self-alignment constraints:}}
\begin{itemize}[leftmargin=1.5em]
\item \textbf{Problem Difficulty \& Thematic Diversity Alignment}: Generate highly complex problems (graduate- or Olympiad-level) covering a wide range of STEM topics. Difficulty is benchmarked against top exams and datasets (GPQA, etc.). Thematic coverage uses role-playing prompts template and a three-tier subject-category-topic framework.

\item \textbf{Logical Consistency Alignment}: Problem-solving must rely solely on rigorous reasoning or systematic derivation, avoiding pattern matching, heuristics, shortcuts, or fabrication. All intermediate steps require justification, preventing logical gaps or errors due to intuition.

\item \textbf{Ground Truth \& Structure Alignment}: Answers must be single, verifiable numerical values (plain numbers, units, ratios, STEM formulas/equations). Avoid hard-to-verify formats (set operations, free text). For multi-solution problems, mandate a specific aggregation (e.g., sum or sum of squares, etc.) for a unique, objectively verifiable answer. Expand beyond single QA to include multi-solution problems (requiring summary values) (e.g., ``calculate total moles of all possible products'').

\item \textbf{Problem Authenticity Alignment}: Problems should be novel, based on authoritative knowledge, but not directly copied. They must be unambiguous, unbiased, accurate, and internally consistent, avoiding nonsensical or hallucinated scenarios.

\item \textbf{Language Consistency Alignment}: The entire generation process (problem statement, reasoning method, solution presentation) must use a single language (e.g., English or Chinese) to prevent multilingual confusion leading to reasoning errors or bad verification cases.

\item \textbf{Problem Structure Consistency Alignment}: Problems must contain only a single primary question, avoiding sub-questions, derivatives, or branching logic that leads to unverifiable cases.

\item \textbf{Modality Consistency Alignment}: Problems must be strictly text-based, describing any necessary complex structures (e.g., chemical molecules, genetic diagrams) textually.

\item \textbf{Formatting Alignment}: Use specific delimiters (e.g., \texttt{<question\_start>}, \texttt{<question\_end>}) for the problem statement and a standardized format (e.g., \texttt{\textbackslash boxed\{\{\$answer\}\}}) for the final answer.
\end{itemize}

\textless{}\textbf{Requirements}\_\textbf{and}\_\textbf{Expectations}\_\textbf{End}\textgreater{}

\end{tcolorbox}
\caption{The \textbf{SHARP} prompt to synthesize high-quality aligned reasoning problems for LRMs reinforcement learning. The colored variables with curly braces in the prompt template are the variables corresponding to the algorithm framework, which will be instantiated with specific values for problem generation.}\label{tab:sharp_prompt}
\end{table}

The raw generated data samples for creating the challenging problems using the \textbf{SHARP} approach can be found via this \href{https://anonymous.4open.science/r/sharp-93F6/sharp_generated_problems_raw.jsonl}{link}.

Next, we conduct a detailed analysis of the coverage of evaluated benchmarks (mainly on GPQA as an example), subject categories related to the dataflow and the difficulty of the STEM challenging problems (mainly on physics, chemistry, and biology) generated by the \textbf{SHARP} method.

\begin{figure}
    \centering
    \includegraphics[width=1\linewidth]{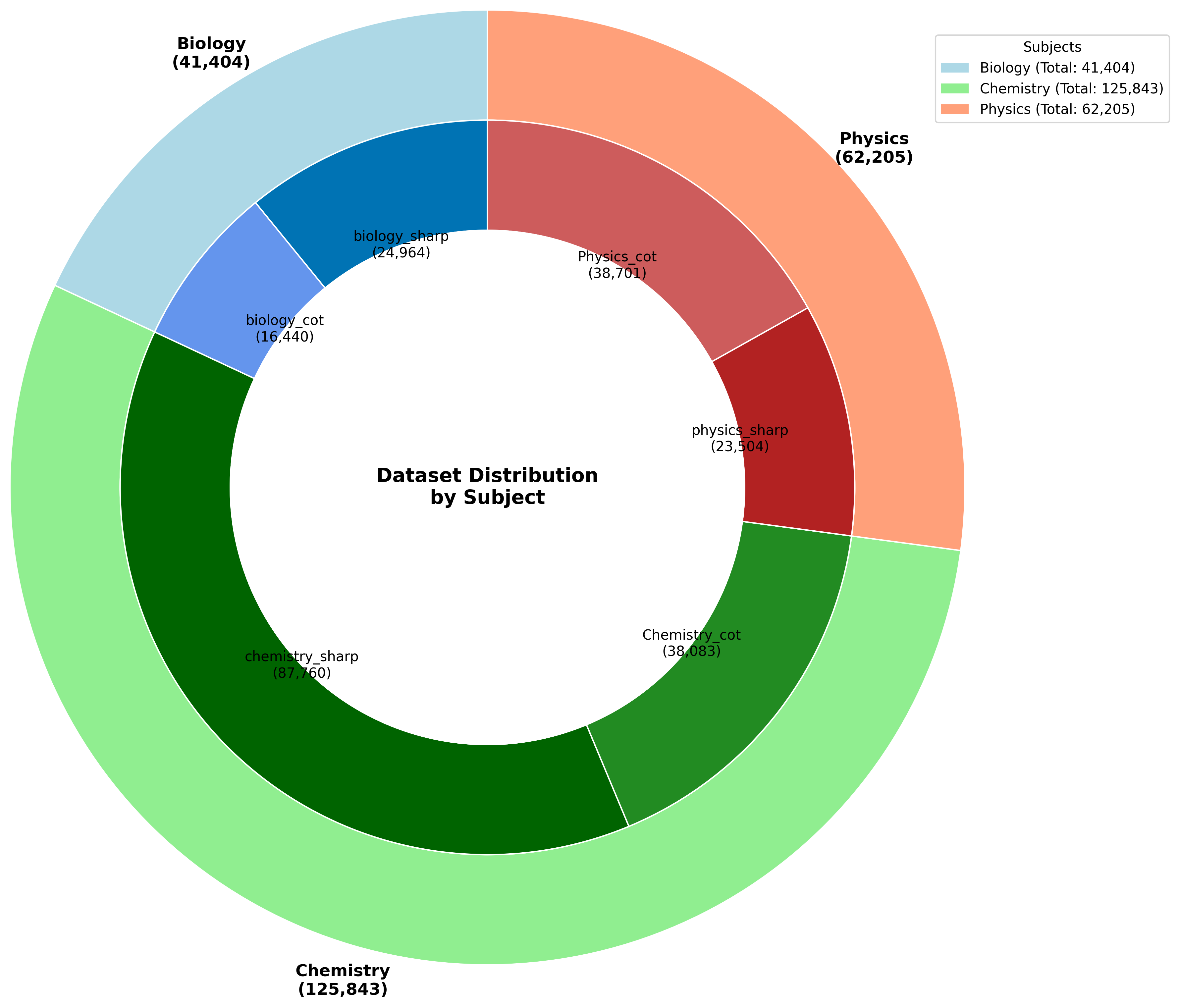}
    \caption{The overall subjects distribution of 229,452 question-answering problems generated by the \textbf{SHARP} approach and the baseline traditional CoT method.}
    \label{fig:stem_dist}
\end{figure}

\subsection{Key Knowledge Point Distribution}
\begin{figure}
    \centering
    \includegraphics[width=1\linewidth]{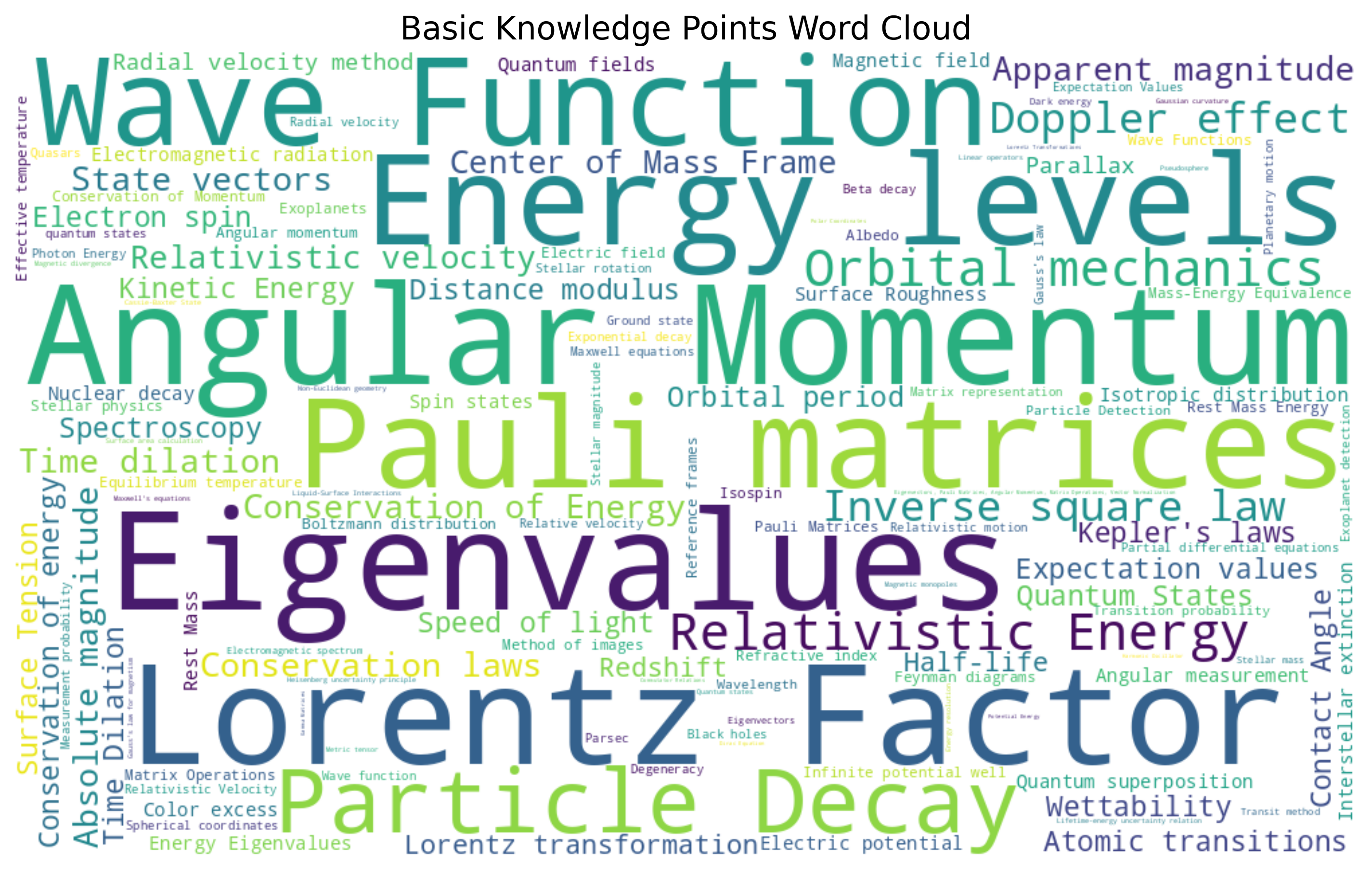}
    \caption{The physics subject distribution of basic knowledge points word cloud of GPQA benchmark.}
    \label{fig:phy_key_dist}
\end{figure}

\begin{figure}
    \centering
    \includegraphics[width=1\linewidth]{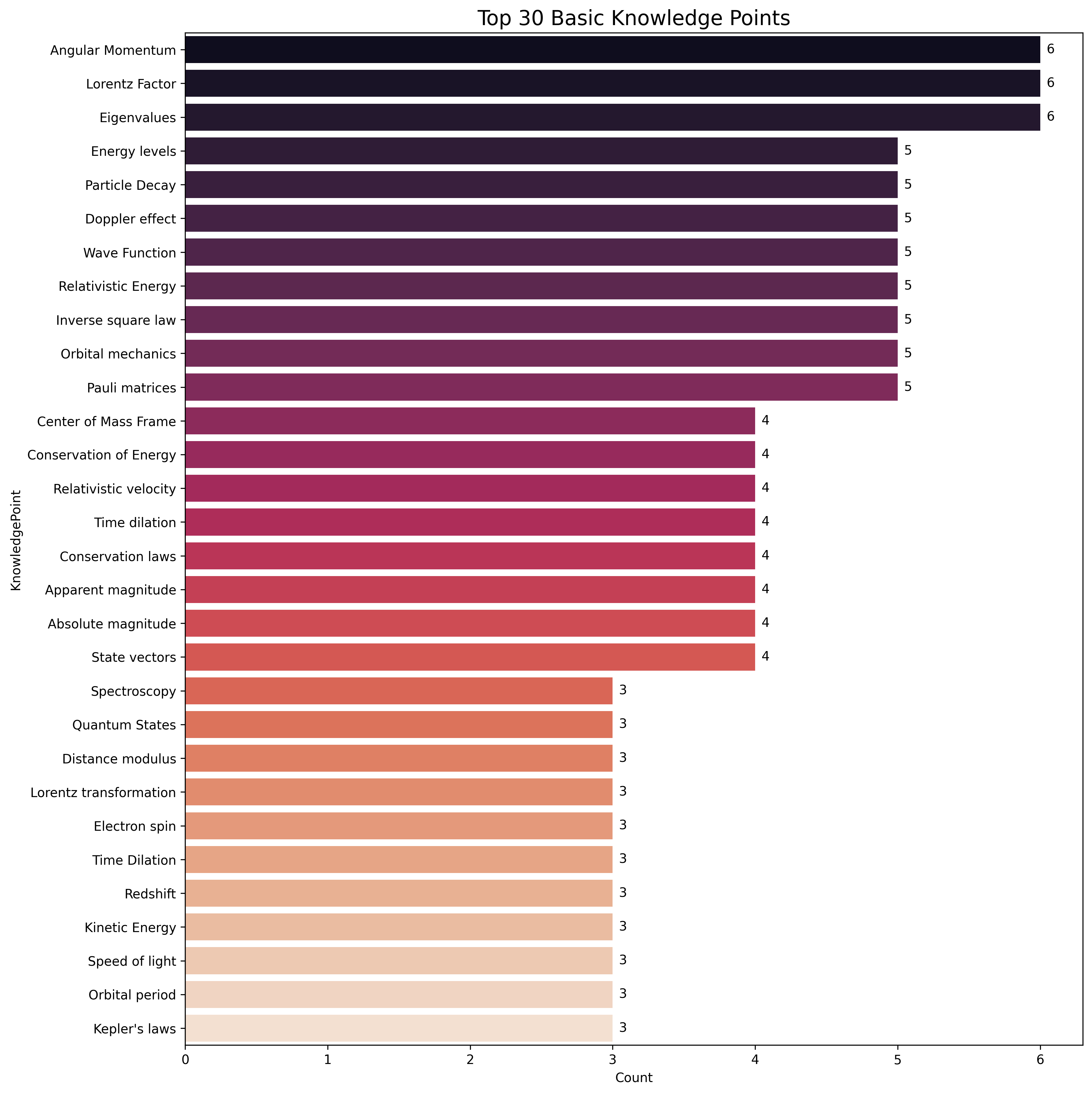}
    \caption{The top 30 basic knowledge points of the physics subject of the GPQA benchmark.}
    \label{fig:phy_basic}
\end{figure}

\begin{figure}
    \centering
    \includegraphics[width=1\linewidth]{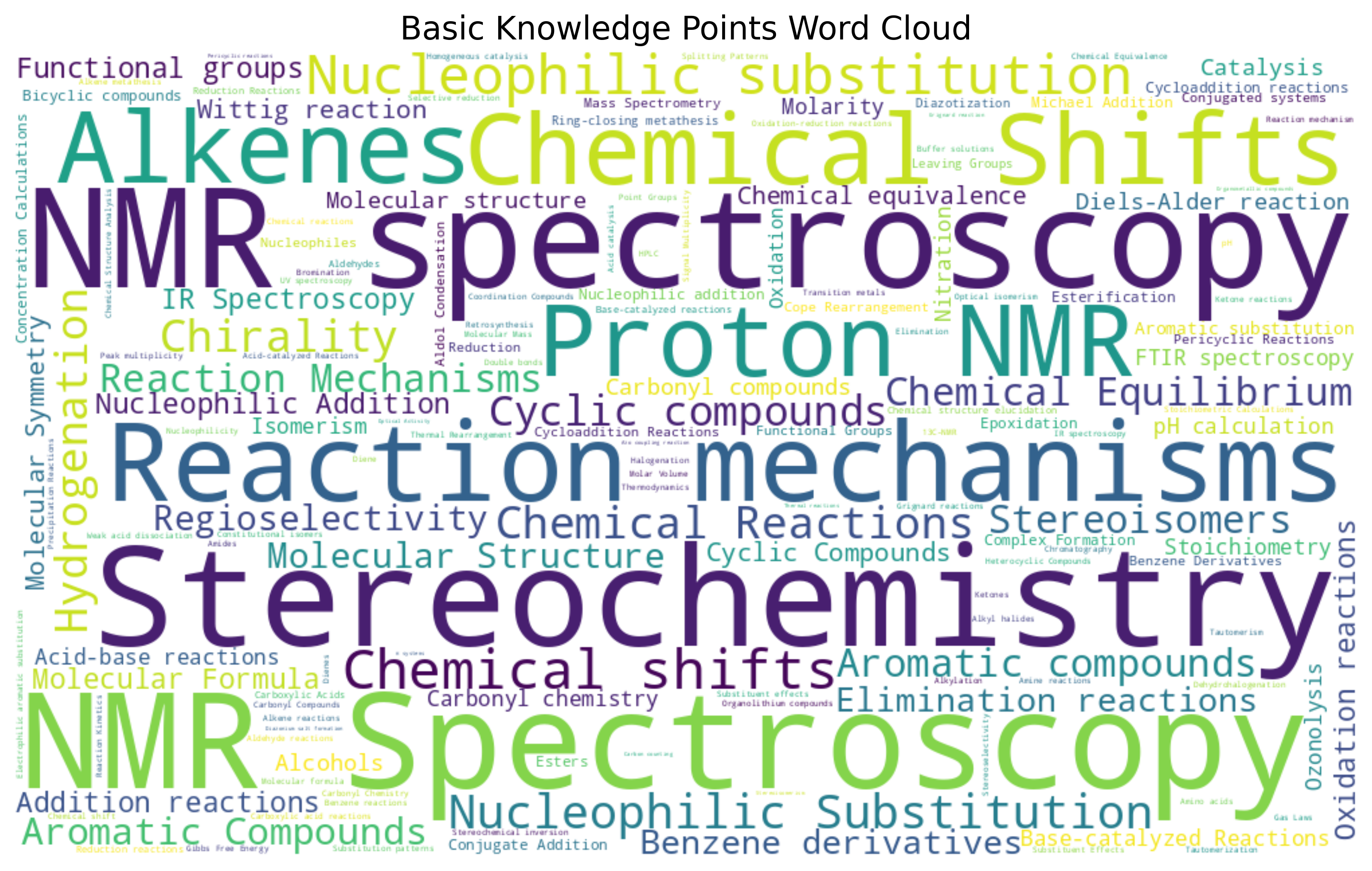}
    \caption{The chemistry subject distribution of basic knowledge points word cloud of GPQA benchmark.}
    \label{fig:chem_key}
\end{figure}

\begin{figure}
    \centering
    \includegraphics[width=1\linewidth]{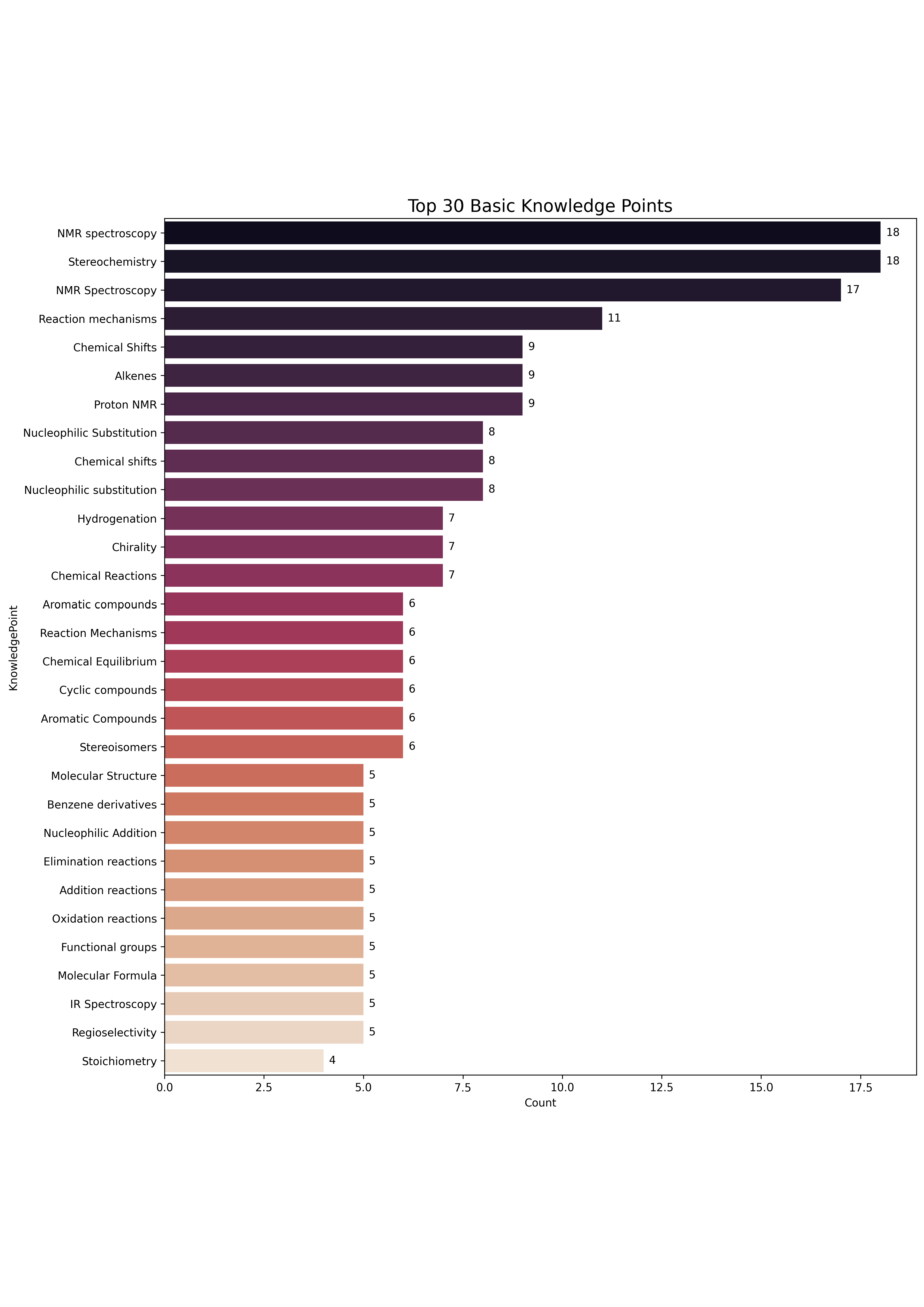}
    \caption{The top 30 basic knowledge points of the chemistry subject of the GPQA benchmark.}
    \label{fig:chem_basic}
\end{figure}

\begin{figure}
    \centering
    \includegraphics[width=1\linewidth]{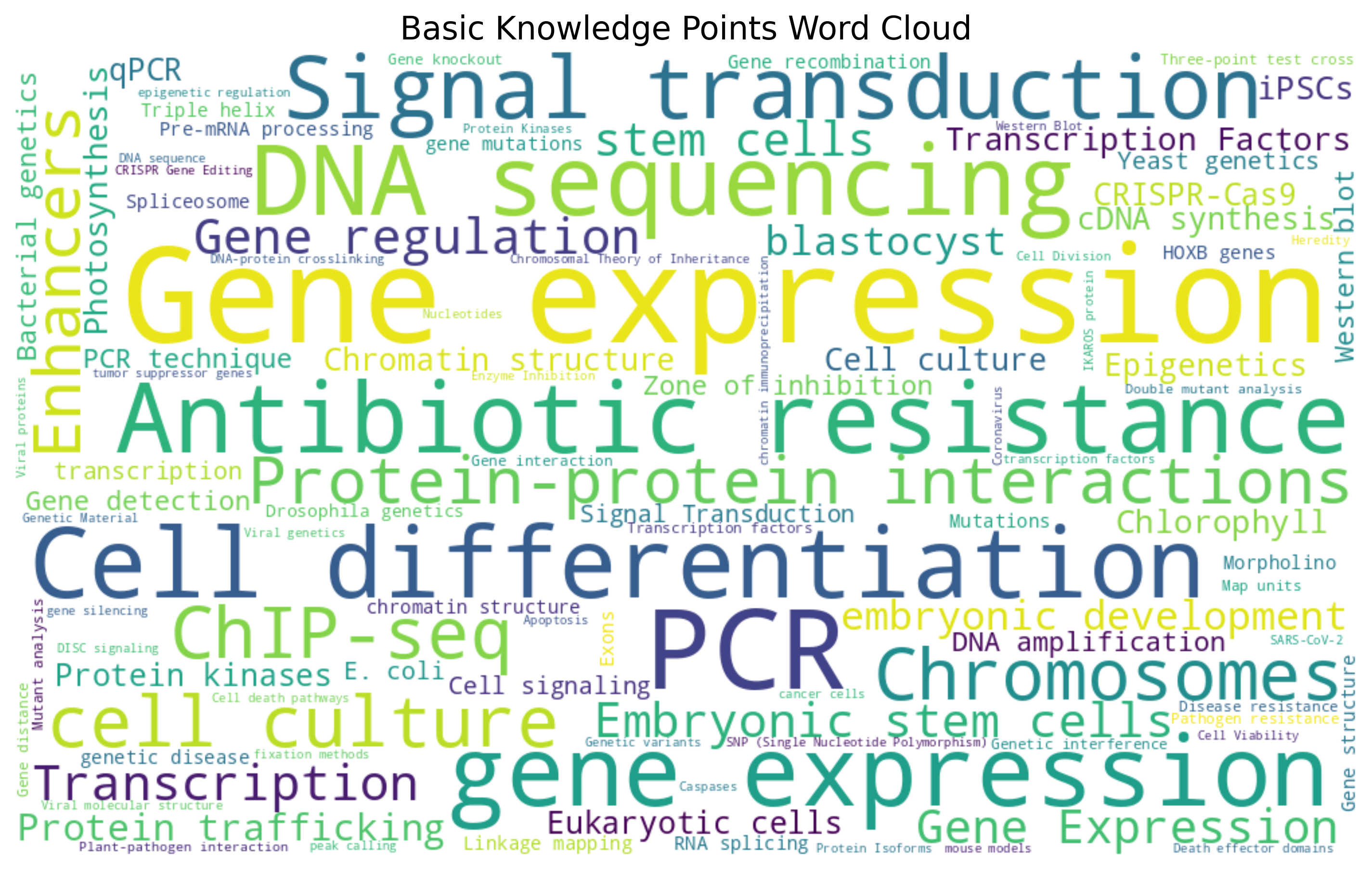}
    \caption{The biology subject distribution of basic knowledge points word cloud of GPQA benchmark.}
    \label{fig:biology_key}
\end{figure}

\begin{figure}
    \centering
    \includegraphics[width=1\linewidth]{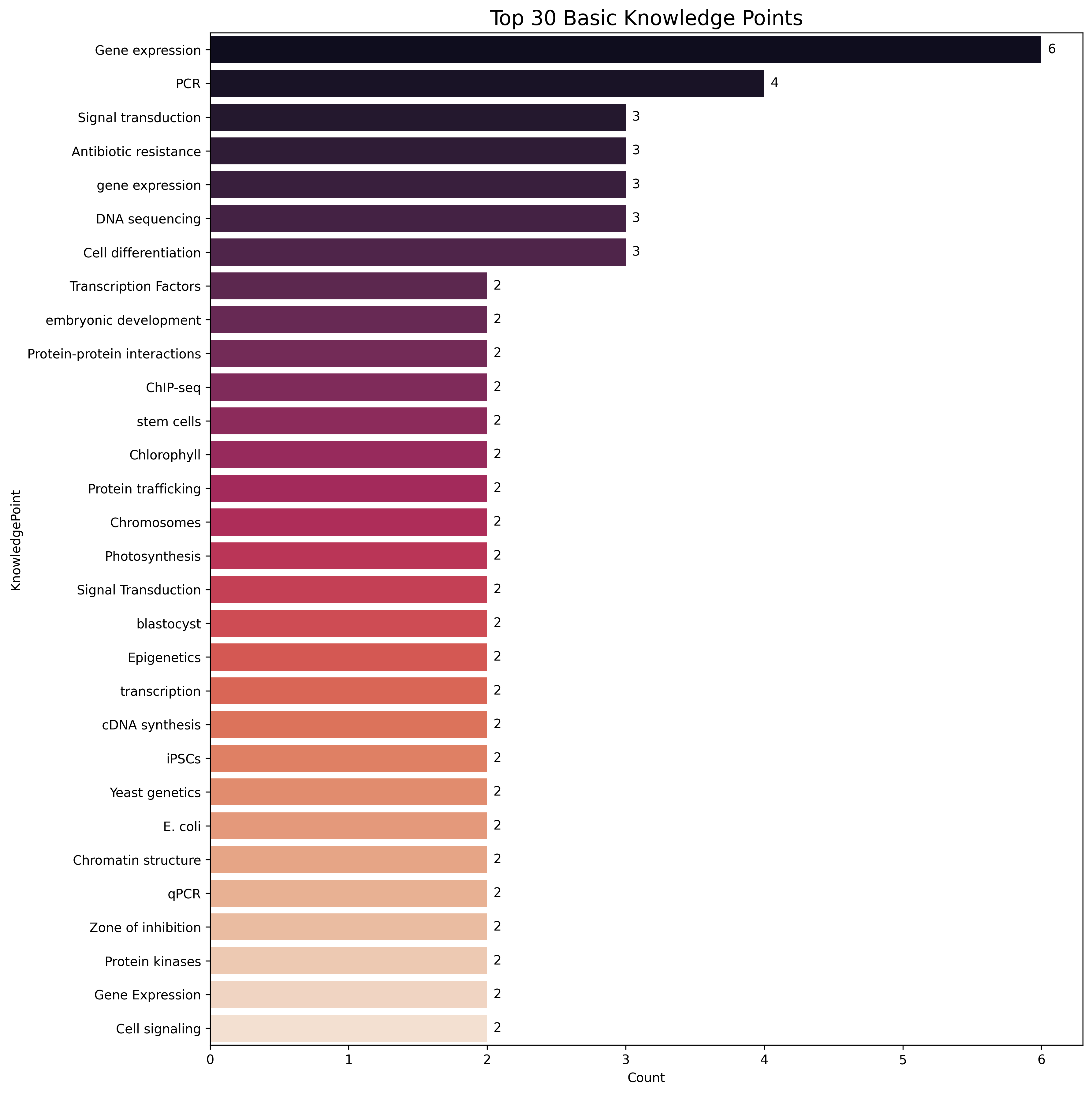}
    \caption{The top 30 basic knowledge points of the biology subject of the GPQA benchmark.}
    \label{fig:biology_basic}
\end{figure}

Through statistical analysis of the distribution of labels and knowledge points, we found that the GPQA benchmark is unevenly distributed, relevant key points distributions shown as in Fig.\ref{fig:phy_key_dist}, \ref{fig:chem_key}, \ref{fig:biology_key} and basic knowledge points shown in Fig.\ref{fig:phy_basic}, \ref{fig:chem_basic}, \ref{fig:biology_basic}. Therefore, in the SHARP data synthesis method, we sampled according to the distribution of disciplines and the corresponding knowledge points to ensure that the synthetic data fully covers the relevant knowledge points of GPQA in terms of distribution.

\subsection{Category Distribution Analysis}
We carefully analyzed the 229,452 STEM question-answering problems generated by the \textbf{SHARP} approach and the baseline traditional CoT method (about 190,000 questions remained after disinfection, deduplication, and decontamination, and pass ratio filtering), with the subject distribution across physics, chemistry, and biology shown in Fig.\ref{fig:stem_dist}. The distribution of subject categories for each subject of physics, chemistry and biology is described below.

\textbf{Physical Category Distribution Analysis}
The category distribution of the synthetic dataset is presented in the following Fig.\ref{fig:phy_dist}.  As observed, the data adheres to a scientifically structured three-level taxonomy.  The first-level categories "Theoretical Physics", "Mechanics", and "Electromagnetism" are the top three categories, encompassing critical second-level disciplines such as Quantum mechanics, Fundamental mechanics, and Electrodynamics. These branches further decompose into highly specialized third-level categories like Theoretical Mechanics, Wave Functions and Schrodinger Equations, Electrostatic Fields, Laws of Thermodynamics—domains particularly effective for evaluating models' reasoning and computational capabilities.
Notably, the dataset maintains substantial diversity despite this concentration, boasting over 200 distinct third-level categories.  This comprehensive coverage across diverse physics domains ensures robust training signals, enabling models to develop balanced proficiency in both dominant and niche scientific reasoning tasks. 

\begin{figure}
    \centering
    \includegraphics[width=1\linewidth]{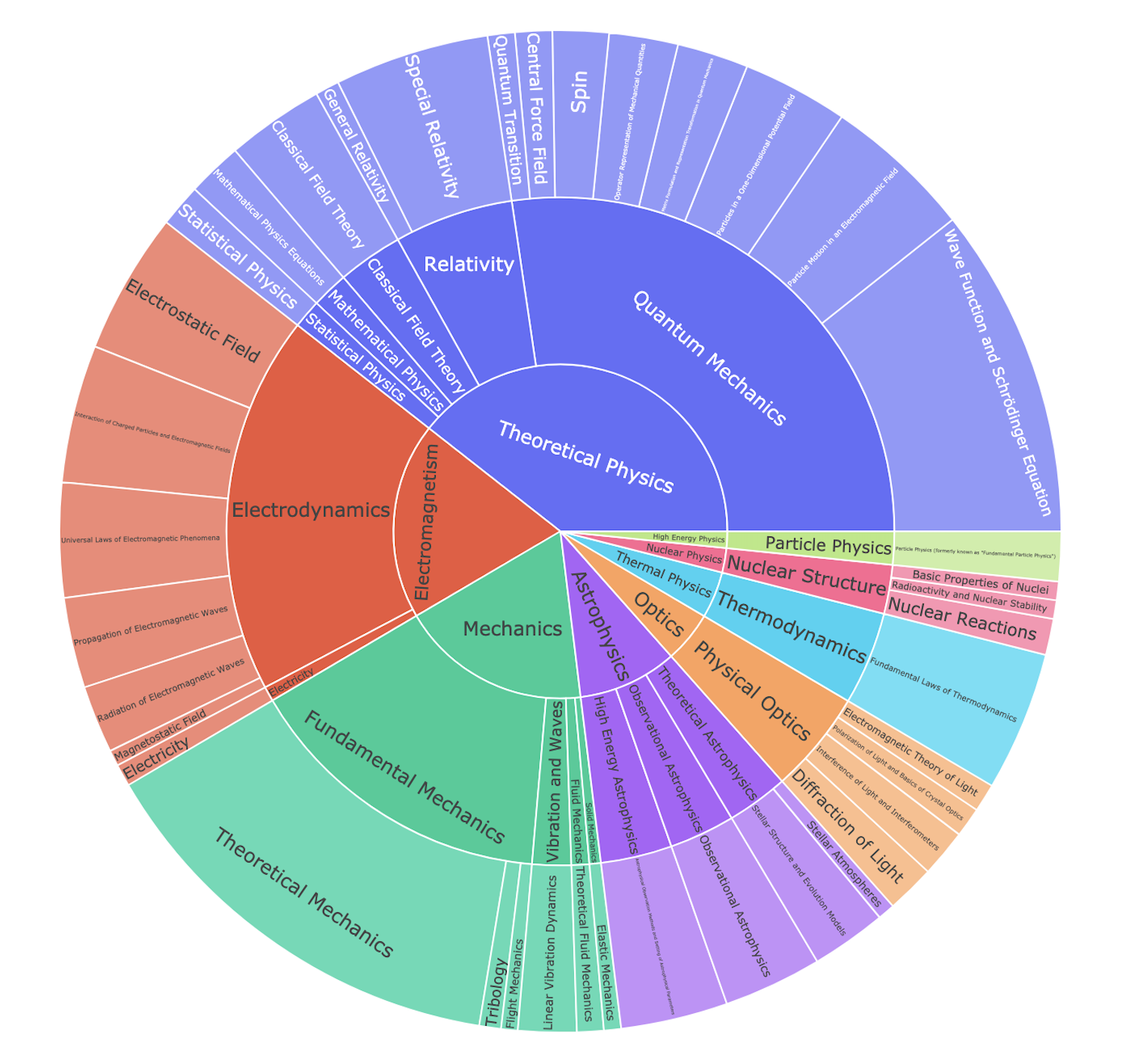}
    \caption{The ``Three-Tier Category'' category distribution of physics subject for problems generated by the \textbf{SHARP} approach.}
    \label{fig:phy_dist}
\end{figure}

\textbf{Biology Category Distribution Analysis}
The category distribution of the synthetic dataset is presented in the following Fig.\ref{fig:biology_dist}.  As observed, the data adheres to a scientifically structured three-level taxonomy.  The first-level category "Fundamental Biology" dominates with over half of the samples, encompassing critical second-level disciplines such as molecular biology, genetics, and cell biology.  These branches further decompose into highly specialized third-level categories like molecular genetics, gene expression, and DNA repair mechanisms—domains particularly effective for evaluating models' reasoning and computational capabilities.
Notably, the dataset maintains substantial diversity despite this concentration, boasting over 100 distinct third-level categories.  This comprehensive coverage across diverse biological domains ensures robust training signals, enabling models to develop balanced proficiency in both dominant and niche scientific reasoning tasks. 

\begin{figure}
    \centering
    \includegraphics[width=1\linewidth]{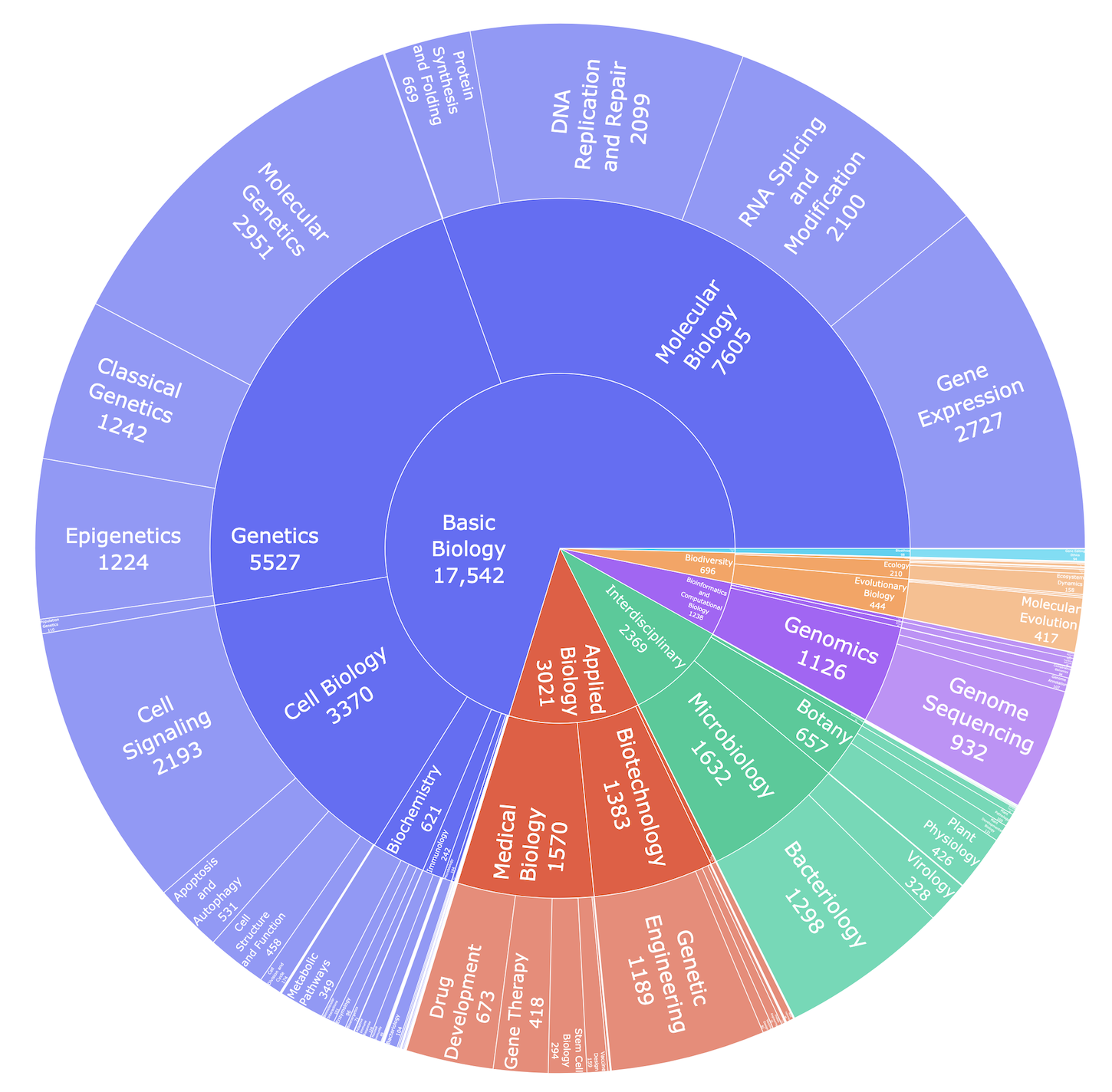}
    \caption{The ``Three-Tier Category'' category distribution of biology subject for problems generated by the \textbf{SHARP} approach.}
    \label{fig:biology_dist}
\end{figure}

\begin{figure}
    \centering
    \includegraphics[width=1\linewidth]{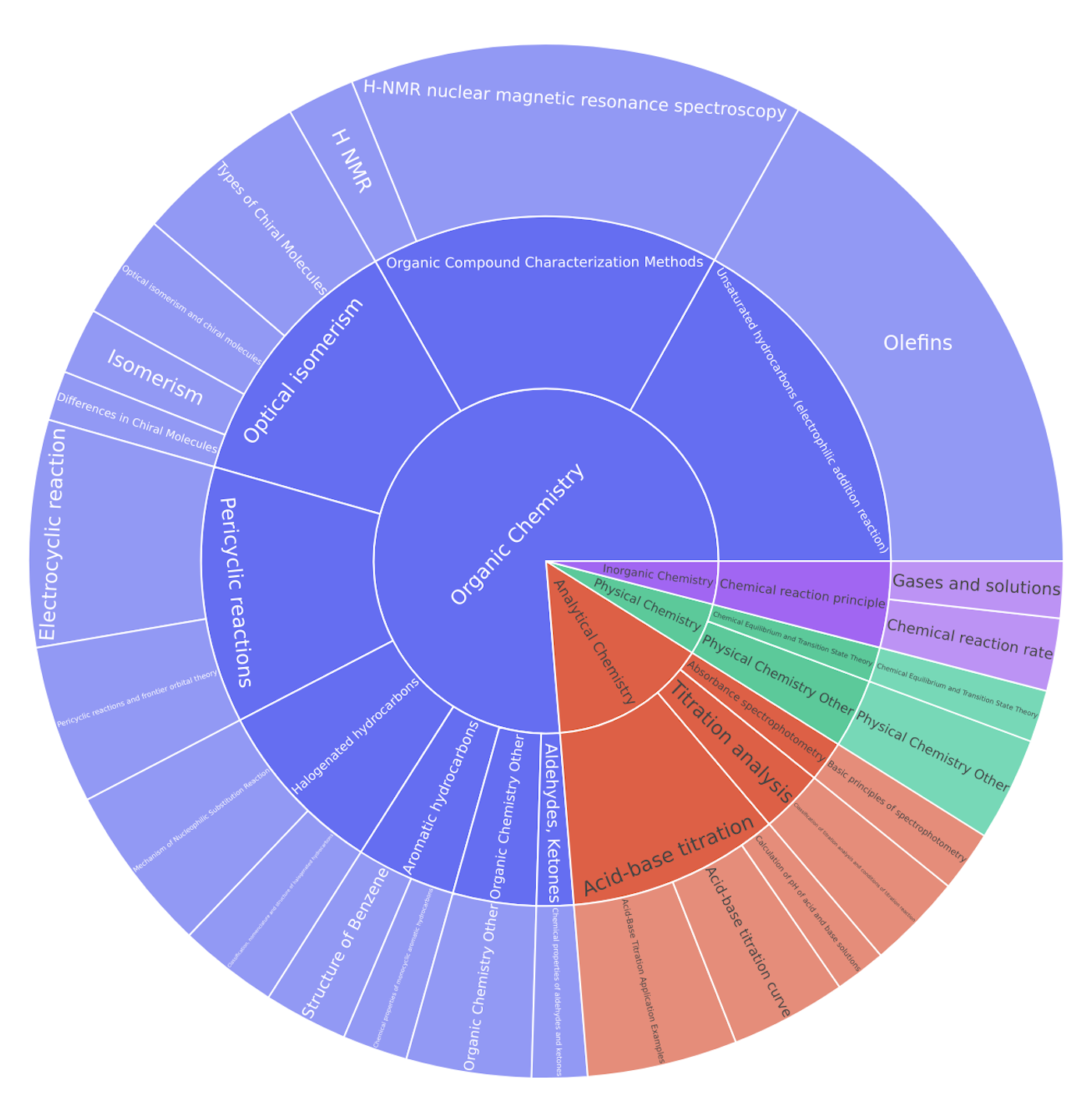}
    \caption{The ``Three-Tier Category'' category distribution of chemistry subject for problems generated by the \textbf{SHARP} approach.}
    \label{fig:chemistry_dist}
\end{figure}

\textbf{Chemistry Category Distribution Analysis}
The analysis of category distribution within the synthetic chemistry dataset is illustrated in the accompanying Fig.\ref{fig:chemistry_dist}. A careful examination reveals that the data adheres to a rigorously structured three-tier category. At the first level, the category of "Organic Chemistry" is predominant, representing more than 75\% of the total samples. This primary category encompasses significant second-level disciplines, including unsaturated hydrocarbons, pericyclic reactions, and the methodologies for characterizing organic compounds. 
These second-level classifications are further delineated into specialized third-level categories, such as olefins, electrocyclic reactions, and H-NMR nuclear magnetic resonance spectroscopy, which are particularly effective in assessing the reasoning and computational capabilities of the models employed. 
Importantly, notwithstanding the dominance of "Organic Chemistry," the dataset exhibits a commendable level of diversity, with over 300 distinct third-level categories represented. This extensive range of coverage across various domains of chemistry fosters robust training signals, thereby facilitating the development of models that exhibit balanced proficiency in both mainstream and niche scientific reasoning tasks.

\subsection{Problem Datasets Difficulty Degree Analysis}
In this section, we present a comprehensive analysis of the difficulty of the problems generated by the \textbf{SHARP} method. 

\begin{figure}
    \centering
    \includegraphics[width=1\linewidth]{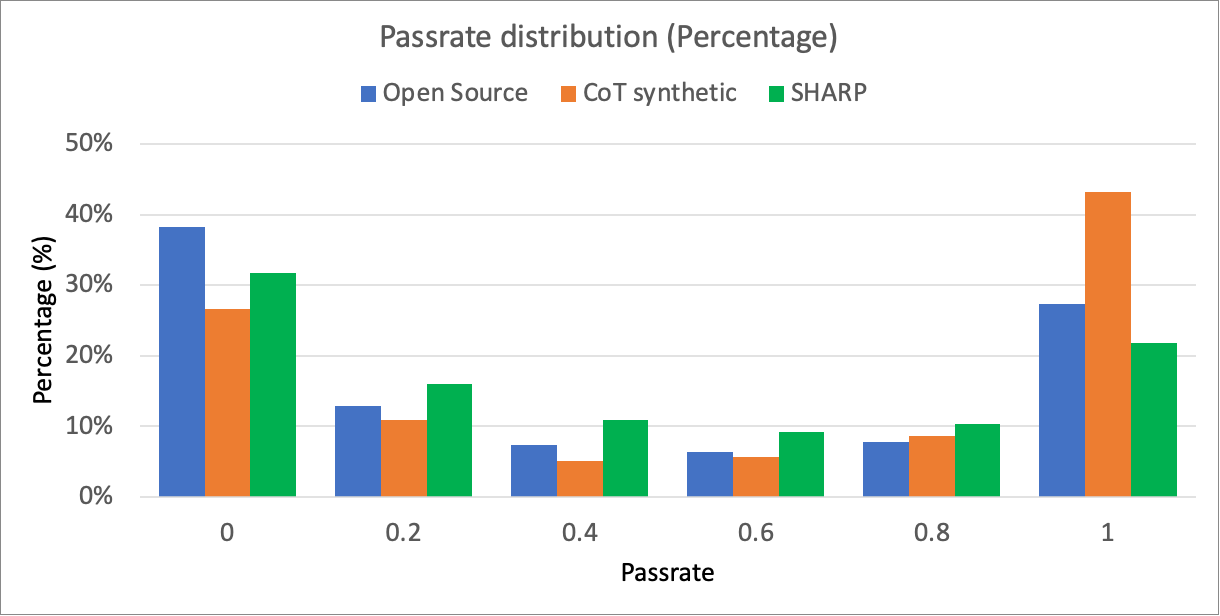}
    \caption{The passrate distribution of the physics problems from open-source , the traditional CoT synthetic, and generated by the \textbf{SHARP} approach.}
    \label{fig:phy_pr_comp}
\end{figure}

\begin{figure}
    \centering
    \includegraphics[width=1\linewidth]{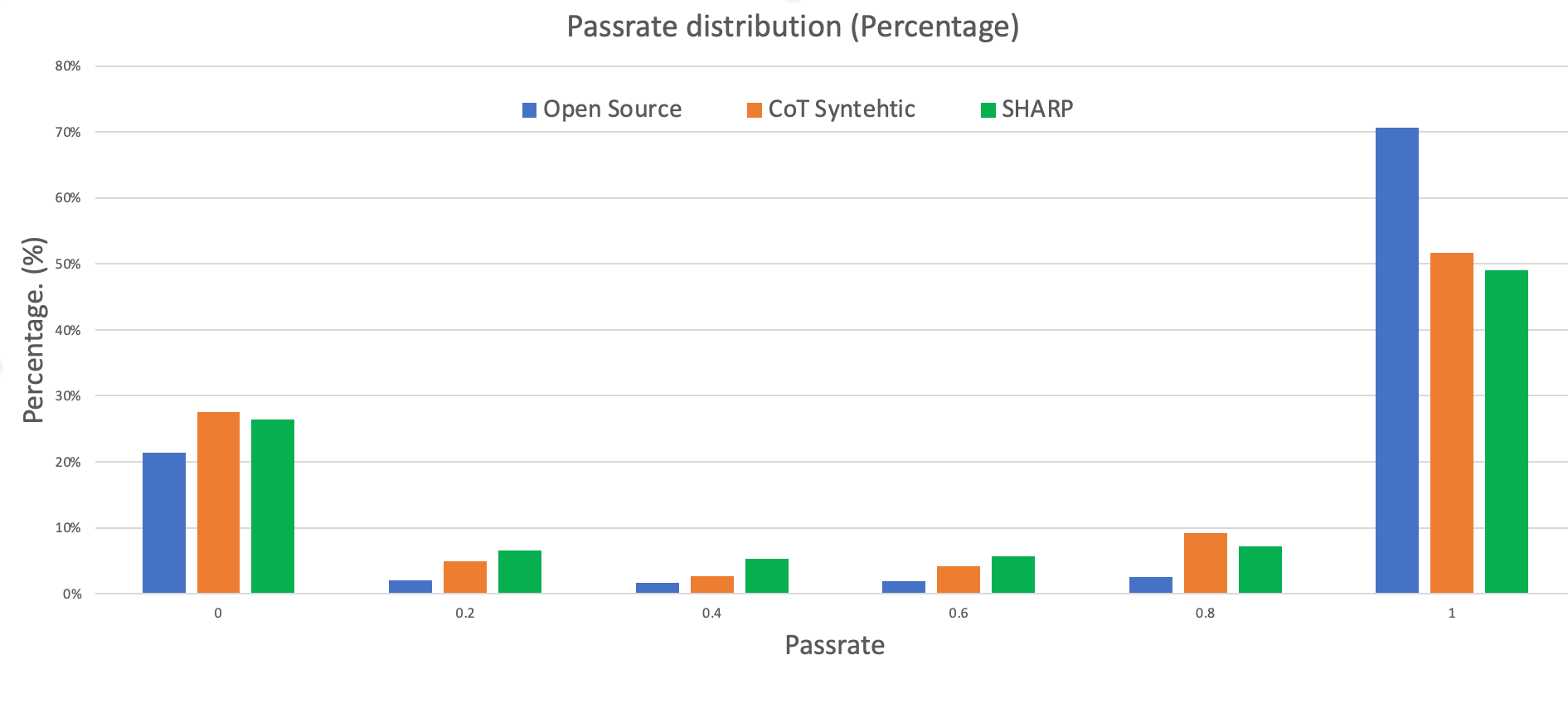}
    \caption{The passrate distribution of the chemistry problems from open-source , the traditional CoT synthetic, and generated by the \textbf{SHARP} approach.}
    \label{fig:chemistry_pr_comp}
\end{figure}

\begin{figure}
    \centering
    \includegraphics[width=1\linewidth]{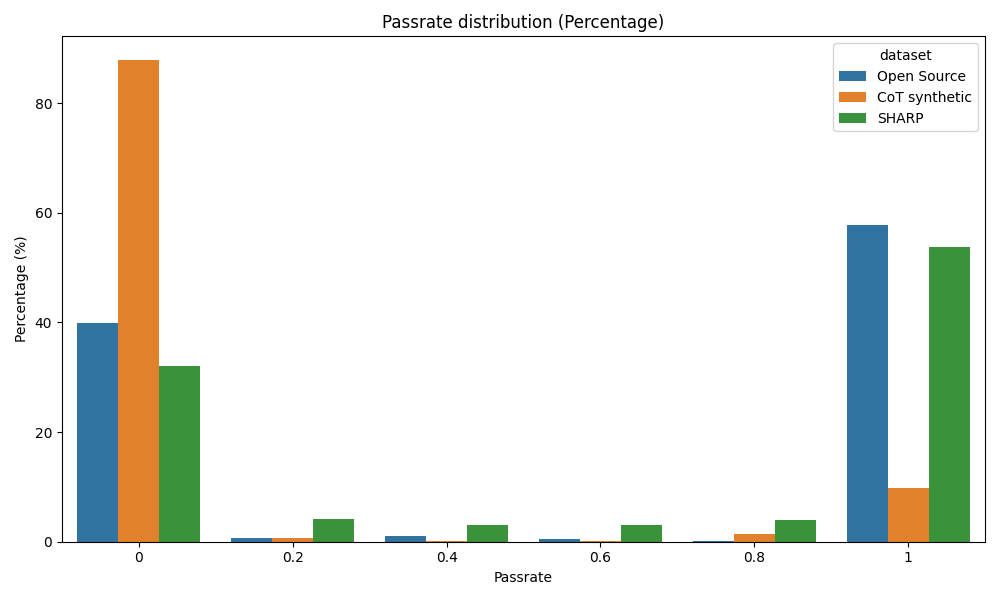}
    \caption{The passrate distribution of the biology problems from open-source , the traditional CoT synthetic, and generated by the \textbf{SHARP} approach.}
    \label{fig:biology_pr_comp}
\end{figure}

\textbf{STEM Pass Rate Distribution Comparison}
Fig.\ref{fig:phy_pr_comp}, \ref{fig:chemistry_pr_comp} and Fig.\ref{fig:biology_pr_comp} illustrate the pass rate distributions of three subjects among three different datasets: the open-source dataset (here refers to the open source data that is mainly based on real data in the industry, with a small amount of open source synthetic data, which is high-quality and challenging after being cleaned, deduplicated and decontaminated), the traditional CoT synthetic dataset, and our \textbf{SHARP} synthetic dataset.  The pass rate is defined as the percentage of correct answers generated by the Qwen2.5-32B-Instruct model over five attempts, where a lower pass rate indicates a higher difficulty level of the question.
As shown in the figure, the difficulty distribution of the \textbf{SHARP} synthetic dataset closely aligns with that of the real-world open-source dataset, making it a viable extension for enhancing the diversity and representativeness of real data.  In contrast, the traditional CoT synthetic dataset exhibits an imbalanced difficulty distribution, with a skewed concentration of either very easy or very challenging questions.
Furthermore, the \textbf{SHARP} dataset demonstrates a well-distributed pass rate across intermediate difficulty levels, providing a multi-level difficulty spectrum for model training.  This balanced distribution enables hierarchical enhancement of the model's reasoning capabilities, ensuring progressive learning and robust performance across tasks of varying complexity.

\begin{figure}
    \centering
    \includegraphics[width=1\linewidth]{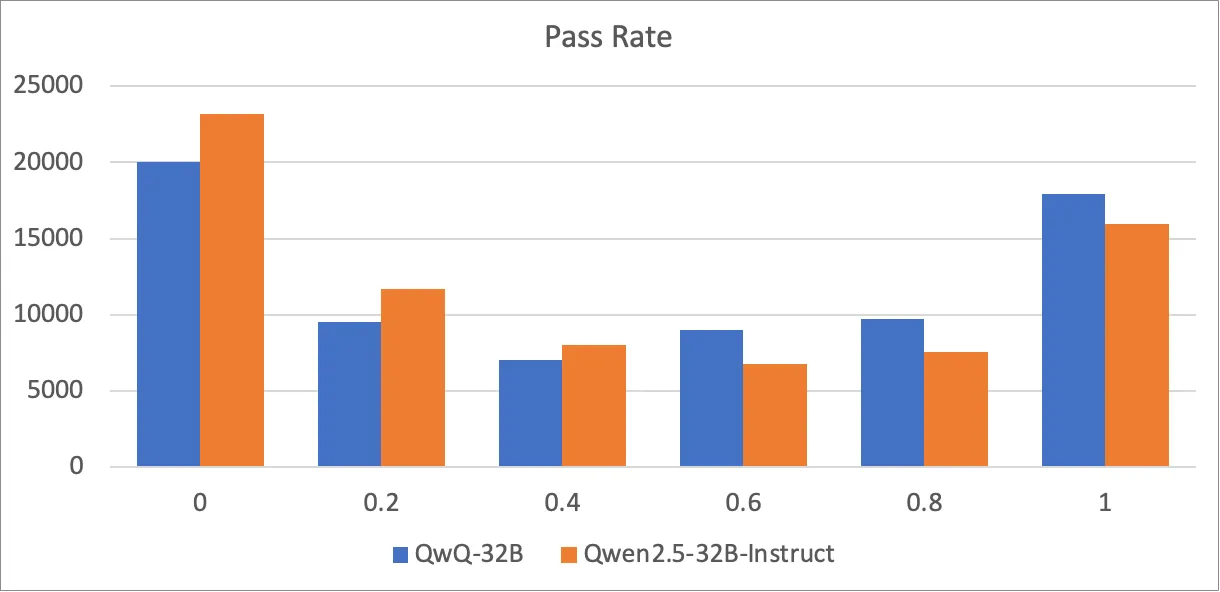}
    \caption{The passrate distribution of physics problems generated by the \textbf{SHARP} approach.}
    \label{fig:phy_pr}
\end{figure}

\textbf{Physics Pass Rate Distributions on the SHARP Dataset}
Fig.\ref{fig:phy_pr} illustrates the pass rate distributions of two models on the \textbf{SHARP} dataset: Qwen2.5-32B-Instruct (based on ten independent responses) and QwQ-32B \citep{qwq32b} (based on five responses). As shown, the QwQ-32B model, which exhibits stronger reasoning capabilities, achieves a significantly higher overall pass rate compared to the Qwen2.5-32B-Instruct model. This is evidenced by a notable reduction in the proportion of questions with a pass rate of 0 and a corresponding increase in the proportion of questions with a pass rate of 1. These results demonstrate the effectiveness of the \textbf{SHARP} dataset in distinguishing the reasoning capabilities of models. By clearly differentiating between models of varying strengths, the \textbf{SHARP} problems dataset can be used to enhance the performance of LLMs' reasoning models in complex tasks.

\begin{figure}
    \centering
    \includegraphics[width=1\linewidth]{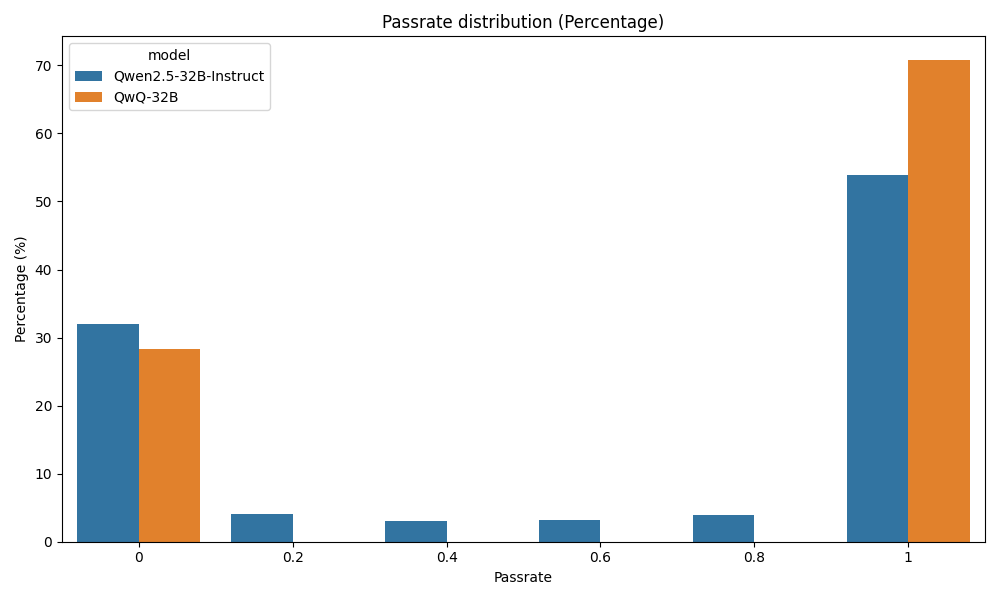}
    \caption{The passrate distribution of biology problems generated by the \textbf{SHARP} approach.}
    \label{fig:biology_pr}
\end{figure}

\textbf{Biology Pass Rate Distributions on the SHARP Dataset}
Fig.\ref{fig:biology_pr} illustrates the pass rate distributions of two models on the \textbf{SHARP} dataset: Qwen2.5-32B-Instruct (based on five independent responses) and QwQ-32B (based on a single response). As shown, the QwQ-32B model, which exhibits stronger reasoning capabilities, achieves a significantly higher overall pass rate compared to the Qwen2.5-32B-Instruct model. This is evidenced by a notable reduction in the proportion of questions with a pass rate of 0 and a corresponding increase in the proportion of questions with a pass rate of 1.
These results demonstrate the effectiveness of the \textbf{SHARP} dataset in distinguishing the reasoning capabilities of models. By clearly differentiating between models of varying strengths, the training dataset generated by \textbf{SHARP} can be used to enhance the performance of reasoning models in complex tasks.

\textbf{Chemistry Pass Rate Distributions on the SHARP Dataset}
The following Fig.\ref{fig:chemistry_pr} presents a comparative analysis of the pass rate distributions for two distinct models evaluated on the \textbf{SHARP} dataset: Qwen2.5-32B-Instruct, which is based on ten independent responses, and QwQ-32B, which relies on five singular responses. The data indicates that the QwQ-32B model, characterized by superior reasoning capabilities, achieves a markedly higher overall pass rate in comparison to the Qwen2.5-32B-Instruct model. This is evidenced by a significant decrease in the proportion of questions that registered a pass rate of 0, alongside a corresponding increase in the proportion of questions attaining a pass rate of 1.
These findings underscore the efficacy of the \textbf{SHARP} dataset in differentiating between the reasoning capabilities of various models. By effectively distinguishing between models with disparate strengths, the \textbf{SHARP} dataset can be used to further enhance the reasoning capabilities of LLMs engaged in complex tasks.

\begin{figure}
    \centering
    \includegraphics[width=1\linewidth]{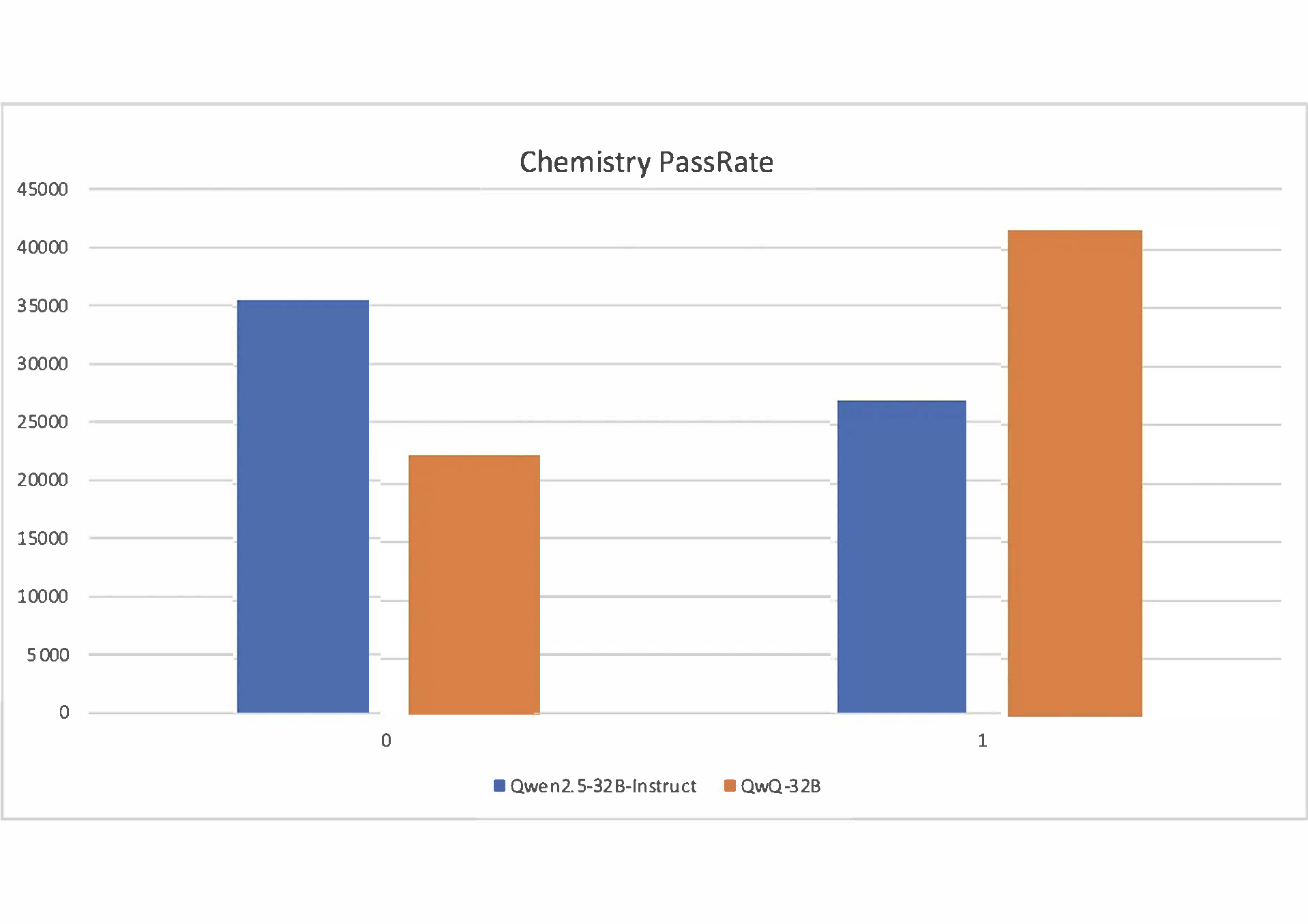}
    \caption{The passrate distribution of chemistry problems generated by the \textbf{SHARP} approach.}
    \label{fig:chemistry_pr}
\end{figure}


\clearpage
\newpage
\section*{NeurIPS Paper Checklist}

\begin{enumerate}
\item {\bf Claims}
    \item[] Question: Do the main claims made in the abstract and introduction accurately reflect the paper's contributions and scope?
    \item[] Answer: \answerYes{} 
    \item[] Justification: The abstract states that this paper introduces \textbf{SHARP}, a unified approach for synthesizing high-quality reasoning problems for LRMs reinforcement learning with verifiable rewards (RLVR). It claims \textbf{SHARP} encompasses self-alignment principles and a three-phase framework. The abstract also claims experiments demonstrate \textbf{SHARP}-augmented training substantially outperforms existing methods. The introduction reiterates these points, highlighting \textbf{SHARP}'s aim to overcome limitations in generating complex STEM reasoning problems and its main components: the \textbf{SHARP} strategy, framework, and implementation. These claims appear to be consistent with the detailed descriptions of the \textbf{SHARP} strategy, framework, implementation, and experimental results presented.

\item {\bf Limitations}
    \item[] Question: Does the paper discuss the limitations of the work performed by the authors?
    \item[] Answer: \answerYes{} 
    \item[] Justification: The paper identifies several limitations in the end. Future work could explore applying this approach to other domains and more complex reasoning tasks, and further optimizing the \textbf{SHARP} approach on various larger-scale RL reasoning foundation models, designing a reward function that weights principles from the \textbf{SHARP} strategy and diving into the distinctions among different subjects, etc. Besides, this paper acknowledges a marginal decrease in GPQA Chemistry performance for the RL-Zero model and attributes it to the nature of chemistry problems and the limitations of unsupervised RL Zero methods without pre-distilled domain-specific priors.

\item {\bf Theory assumptions and proofs}
    \item[] Question: For each theoretical result, does the paper provide the full set of assumptions and a complete (and correct) proof?
    \item[] Answer: \answerNA{} 
    \item[] Justification: The paper introduces a new approach (\textbf{SHARP}) and a framework, supported by experimental results. It does not appear to present new theoretical results in the form of theorems or mathematical proofs that would require a separate section for assumptions and proofs.

    \item {\bf Experimental result reproducibility}
    \item[] Question: Does the paper fully disclose all the information needed to reproduce the main experimental results of the paper to the extent that it affects the main claims and/or conclusions of the paper (regardless of whether the code and data are provided or not)?
    \item[] Answer: \answerYes{} 
    \item[] Justification: The paper describes the training data, including the baseline dataset and the \textbf{SHARP}-generated dataset of 190,000 samples. It specifies the comparison models used for distillation and RL Zero training. It mentions that training details for distillation involved standard procedures, and for \textbf{SHARP}-RL Zero training, the GRPO algorithm was used with a rule-based reward function, with hyperparameters and computational resources detailed in the appendix. The evaluation metrics are centered on the GPQA STEM reasoning benchmark using accuracy metrics like pass@k. The appendix also provides further details, including ablation studies and analysis of the generated dataset's distribution.

\item {\bf Open access to data and code}
    \item[] Question: Does the paper provide open access to the data and code, with sufficient instructions to faithfully reproduce the main experimental results, as described in supplemental material?
    \item[] Answer: \answerYes{} 
    \item[] Justification: Essential \textbf{SHARP} strategy prompts template and specific problems for different subjects are included in the appendix. Moreover, we will open-source all necessary codes and related data for industry use during the review period.

\item {\bf Experimental setting/details}
    \item[] Question: Does the paper specify all the training and test details (e.g., data splits, hyperparameters, how they were chosen, type of optimizer, etc.) necessary to understand the results?
    \item[] Answer: \answerYes{} 
    \item[] Justification: The paper specifies the training data used (baseline CoT samples and 190,000 SHARP-generated samples). It names the models used for comparison. For SHARP-RL Zero training, the GRPO algorithm and a rule-based reward function hyperparameters, and computational resources are detailed in the appendix. The evaluation benchmark (GPQA) and metrics (accuracy, pass@k) are also clearly stated. The appendix further details some of these aspects.

\item {\bf Experiment statistical significance}
    \item[] Question: Does the paper report error bars suitably and correctly defined or other appropriate information about the statistical significance of the experiments?
    \item[] Answer: \answerNo{} 
    \item[] Justification: The tables presenting the main experimental results (Table 1 and Table 2) show performance scores (e.g., GPQA Diamond scores, scores for Physics, Chemistry, Biology) but do not include error bars, confidence intervals, or mention statistical significance tests. If needed, we will include them in the camera-ready version.

\item {\bf Experiments compute resources}
    \item[] Question: For each experiment, does the paper provide sufficient information on the computer resources (type of compute workers, memory, time of execution) needed to reproduce the experiments?
    \item[] Answer: \answerYes{} 
    \item[] Justification: The computational resources are detailed in the appendix.
    
\item {\bf Code of ethics}
    \item[] Question: Does the research conducted in the paper conform, in every respect, with the NeurIPS Code of Ethics \url{https://neurips.cc/public/EthicsGuidelines}?
    \item[] Answer: \answerYes{} 
    \item[] Justification: The research involves algorithmic development and evaluation on standard optimization benchmarks. It does not involve human subjects or obviously ethically sensitive applications, and we assume it conforms to the NeurIPS Code of Ethics.

\item {\bf Broader impacts}
    \item[] Question: Does the paper discuss both potential positive societal impacts and negative societal impacts of the work performed?
    \item[] Answer: \answerNo{} 
    \item[] Justification: The paper focuses on the technical contributions of the \textbf{SHARP} approach in enhancing LRM reasoning capabilities, particularly in STEM domains. It discusses the potential to push LRM performance closer to expert-level proficiency and superintelligence in STEM. However, it does not contain a dedicated section or explicit discussion on broader societal impacts, either positive or negative, beyond the advancement of AI reasoning capabilities.
    
\item {\bf Safeguards}
    \item[] Question: Does the paper describe safeguards that have been put in place for responsible release of data or models that have a high risk for misuse (e.g., pretrained language models, image generators, or scraped datasets)?
    \item[] Answer: \answerNA{} 
    \item[] Justification:  The paper focuses on generating challenging STEM problems and does not explicitly state it is releasing a pre-trained language model or a dataset scraped from sources that would pose a high risk for misuse in the sense described by the guidelines (e.g., generating deepfakes). The generated data consists of STEM problems. While advanced AI models could have dual-use potential, the paper does not discuss releasing models or data in a way that would necessitate specific safeguards as outlined.

\item {\bf Licenses for existing assets}
    \item[] Question: Are the creators or original owners of assets (e.g., code, data, models), used in the paper, properly credited and are the license and terms of use explicitly mentioned and properly respected?
    \item[] Answer: \answerNo{} 
    \item[] Justification: The paper properly cites the sources for existing assets like baseline models (e.g., DeepSeek R1, Qwen models) and benchmarks like GPQA. However, the specific licenses and terms of use for these assets are not explicitly mentioned in the paper text or the appendix.

\item {\bf New assets}
    \item[] Question: Are new assets introduced in the paper well documented and is the documentation provided alongside the assets?
    \item[] Answer: \answerYes{} 
    \item[] Justification: The main new asset introduced is the \textbf{SHARP} methodology and the dataset of 190,000 STEM problems generated using this methodology. The paper provides extensive documentation on the \textbf{SHARP} strategy (Algorithm 1), the \textbf{SHARP} framework (Alignment, Instantiation, Inference phases, Three-Tier Category knowledge structure), and the \textbf{SHARP} implementation. Appendix B provides a detailed analysis of the 190,000 \textbf{SHARP}-generated samples, including distributions across STEM subcategories and pass rate analyses. This constitutes detailed documentation of the new asset (the problem generation methodology and the resulting dataset characteristics). 

\item {\bf Crowdsourcing and research with human subjects}
    \item[] Question: For crowdsourcing experiments and research with human subjects, does the paper include the full text of instructions given to participants and screenshots, if applicable, as well as details about compensation (if any)? 
    \item[] Answer: \answerNA{} 
    \item[] Justification: The paper does not describe any crowdsourcing experiments or research involving human subjects as participants in studies. The process involves using LLMs to generate and verify problems, and then training other LLMs.

\item {\bf Institutional review board (IRB) approvals or equivalent for research with human subjects}
    \item[] Question: Does the paper describe potential risks incurred by study participants, whether such risks were disclosed to the subjects, and whether Institutional Review Board (IRB) approvals (or an equivalent approval/review based on the requirements of your country or institution) were obtained?
    \item[] Answer: \answerNA{} 
    \item[] Justification:  As the paper does not involve research with human subjects, IRB approval is not applicable.

\item {\bf Declaration of LLM usage}
    \item[] Question: Does the paper describe the usage of LLMs if it is an important, original, or non-standard component of the core methods in this research? Note that if the LLM is used only for writing, editing, or formatting purposes and does not impact the core methodology, scientific rigorousness, or originality of the research, declaration is not required.
    \item[] Answer: \answerYes{} 
    \item[] Justification: The core methodology of \textbf{SHARP} heavily involves the use of Large Reasoning Models (LRMs). The paper states, "We implement SHARP by leveraging a state-of-the-art LRM to infer and verify challenging STEM questions" and "Our proposed \textbf{SHARP} approach aims to systematically generate high-quality, complex STEM reasoning samples by guiding a state-of-the-art LRM (such as DeepSeek R1) instance-alignment reasoning inference through the \textbf{SHARP} framework". The use of LRMs is central to the problem generation and refinement process, making it an important and original component of the research.

\end{enumerate}

\end{document}